\documentclass{article}
\PassOptionsToPackage{numbers, compress}{natbib}
\usepackage[preprint]{neurips_2024} %
\usepackage[utf8]{inputenc}
\usepackage[T1]{fontenc}
\usepackage{amsfonts}
\usepackage{amsmath}
\usepackage{enumitem}
\usepackage{booktabs}
\usepackage[dvipsnames]{xcolor}         %
\usepackage{graphicx}
\usepackage{hyperref}
\usepackage{lipsum}
\usepackage{nicefrac}
\usepackage{makecell}
\usepackage{microtype}
\usepackage{multirow}
\usepackage{textcomp}
\usepackage{url}
\usepackage{wrapfig}
\usepackage{xcolor}
\usepackage{xspace}
\usepackage[capitalize,nameinlink]{cleveref}
\newcommand{\benchmark}{\textsc{LOFT}}
\newcommand{\Method}{Corpus-in-Context Prompting}
\newcommand{\method}{CiC}
\newcommand{\lclms}{LCLMs}
\newcommand{\lclm}{LCLM}
\newcommand{\dialogue}{DialogRE}
\newcommand{\LiclBench}{LongICLBench}

\newcommand{\eat}[1]{}

\usepackage{pifont}
\newcommand{\cmark}{\ding{51}}
\newcommand{\xmark}{\ding{55}}
\newcolumntype{M}[1]{>{\centering\arraybackslash}m{#1}}

\crefformat{section}{\S#2#1#3} %
\crefformat{subsection}{\S#2#1#3}
\crefformat{subsubsection}{\S#2#1#3}

\newcommand{\correct}{{\color{green}{(\cmark)}}}
\newcommand{\wrong}{{\color{red}{(\xmark)}}}

\title{Can Long-Context Language Models Subsume\\Retrieval, RAG, SQL, and More?}

\newcommand*\samethanks[1][\value{footnote}]{\footnotemark[#1]}

\author{%
  \textbf{Jinhyuk Lee}\thanks{Lead contributors.} \quad
  \textbf{Anthony Chen}\samethanks[1] \quad
  \textbf{Zhuyun Dai}\samethanks[1] \\[.3em]
  \textbf{Dheeru Dua} \quad
  \textbf{Devendra Singh Sachan} \quad
  \textbf{Michael Boratko} \quad
  \textbf{Yi Luan} \\[.3em]
  \textbf{Sébastien M. R. Arnold} \quad
  \textbf{Vincent Perot} \quad
  \textbf{Siddharth Dalmia} \quad
  \textbf{Hexiang Hu} \\[.3em]
  \textbf{Xudong Lin} \quad
  \textbf{Panupong Pasupat} \quad
  \textbf{Aida Amini} \quad
  \textbf{Jeremy R. Cole} \\[.3em]
  \textbf{Sebastian Riedel} \quad \textbf{Iftekhar Naim} \quad \textbf{Ming-Wei Chang} \quad \textbf{Kelvin Guu}\\[.7em]
  Google DeepMind
  \vspace{-3mm}
}

\begin{document}
\maketitle
\setcounter{footnote}{0}
\begin{abstract}
    Long-context language models (\lclms) have the potential to revolutionize our approach to tasks traditionally reliant on external tools like retrieval systems or databases.
    Leveraging \lclms{}' ability to natively ingest and process entire corpora of information  offers numerous advantages.
    It enhances user-friendliness by eliminating the need for specialized knowledge of tools, provides robust end-to-end modeling that minimizes cascading errors in complex pipelines, and allows for the application of sophisticated prompting techniques across the entire system.
    To assess this paradigm shift, we introduce \benchmark{}, a benchmark of real-world tasks requiring context up to millions of tokens designed to evaluate \lclms{}' performance on in-context retrieval and reasoning.
   Our findings reveal LCLMs' surprising ability to rival state-of-the-art retrieval and RAG systems, despite never having been explicitly trained for these tasks.
    However, \lclms{} still face challenges in areas like compositional reasoning that are required in SQL-like tasks.
    Notably, prompting strategies significantly influence performance, emphasizing the need for continued research as context lengths grow. 
    Overall, \benchmark{} provides a rigorous testing ground for \lclms{}, showcasing their potential to supplant existing paradigms and tackle novel tasks as model capabilities scale.\footnote{The \benchmark{} benchmark is available at \url{https://github.com/google-deepmind/loft}.}
\begin{figure}[t]
\centering
\includegraphics[width=0.99\columnwidth]{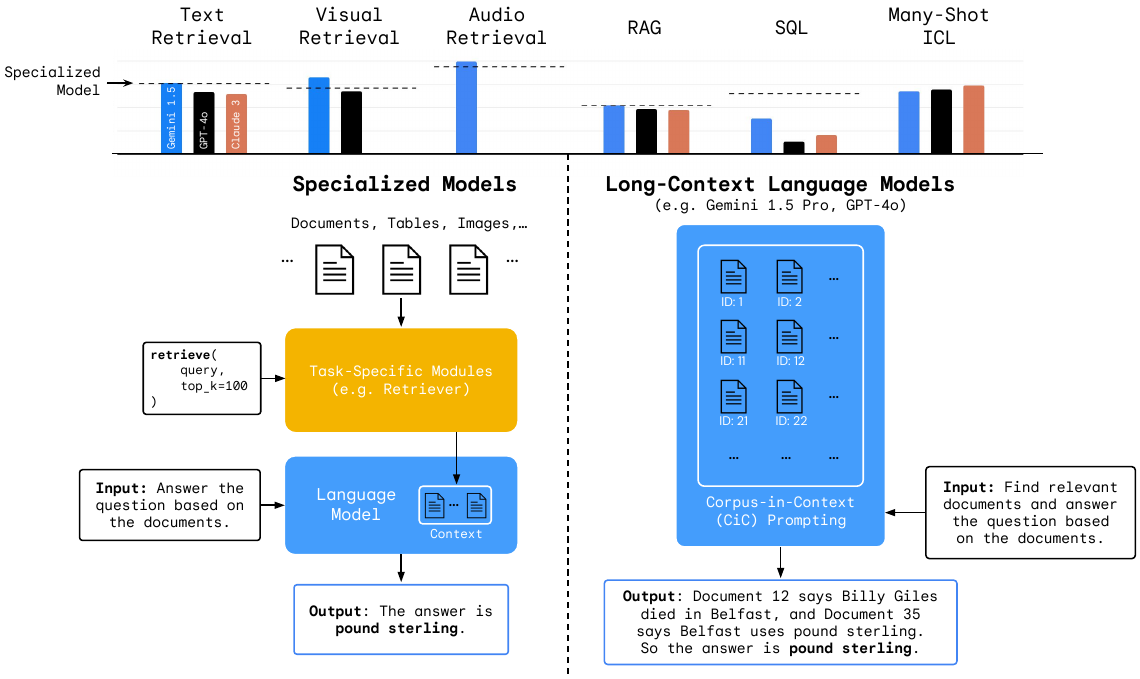}
\vspace{-3pt}
\caption{
    \label{fig:overview}
    An overview of the \benchmark{} benchmark, made of six tasks which measure \lclms{}' ability to do in-context retrieval, reasoning, and many-shot learning on corpora up to millions of tokens.
    We compare the performance of \lclms{} against specialized models (\textit{e.g.,} CLIP for visual retrieval), which often rely on complex task-specific fine-tuning or pipelining.
    Unlike specialized models, we show how \lclms{} can simplify various tasks through Corpus-in-Context Prompting (\Cref{section:prompting}).
}
\vspace{-5pt}
\end{figure}

\end{abstract}

\section{Introduction}\label{section:introduction}
    Long-context language models (\lclms)~\citep{beltagy2020longformer,guo2022longt5,Achiam2023GPT4TR,TheC3,Reid2024Gemini1U} hold the promise of reshaping artificial intelligence by enabling entirely new tasks and applications while eliminating the reliance on tools and complex pipelines previously necessary due to context length limitations~\citep{Guu2020REALMRL, Lewis2020RetrievalAugmentedGF}.
    By consolidating complex pipelines into a unified model, \lclms{} ameliorate issues like cascading errors~\citep{Barnett2024SevenFP} and cumbersome optimization~\citep{Lee2021YouON, Xiong2020ApproximateNN}, offering a streamlined end-to-end approach to model development.
    Moreover, techniques such as adding instructions \citep{Kojima2022LargeLM, Wei2021FinetunedLM, Chung2022ScalingIL}, incorporating few-shot examples~\citep{Brown2020LanguageMA}, and leveraging demonstrations via chain-of-thought prompting \citep{Nye2021ShowYW, Wei2022ChainOT} can be seamlessly integrated to optimize \lclms{} for the task at hand.

    However, realizing the full potential of \lclms{} requires rigorous evaluation on truly long-context tasks useful in real-world applications.
    Existing benchmarks fall short in this regard, relying on synthetic tasks like the popular ``needle-in-haystack'' \citep{KamradtNeedle, Levy2024SameTM} or fixed-length datasets that fail to keep pace with the evolving definition of ``long-context'' \citep{Bai2023LongBenchAB}.
    Critically, existing evaluations do not adequately stress-test \lclms{} on any paradigm-shifting tasks.

    To address this, we introduce the \textbf{Lo}ng-Context \textbf{F}ron\textbf{t}iers~(\benchmark{}) benchmark, a suite of six tasks  consisting of 35 datasets which span text, visual, and audio modalities, to push \lclms{} to their limits and gauge their real-world impact.
    Unlike previous benchmarks, \benchmark{} allows for automatic creation of increasing context lengths. While the current version extends to one million tokens, it can easily be extended further to tens of millions, ensuring rigorous evaluation as \lclms{} continue to scale.
    \benchmark{} focuses on the following areas where \lclms{} have the potential for disruption:
    
    \begin{itemize}[leftmargin=*]
        \item \textbf{Retrieval}:
            \lclms{} can directly ingest and retrieve information from a corpus, eliminating the need for separate dual-encoder models~\citep{Karpukhin2020DensePR,ni2022large,Lee2024GeckoVT,clip}. This addresses long-standing challenges in retrieval systems such as multi-hop reasoning, instruction following, and few-shot task adaptation. 
            We assess retrieval performance across text, visual, and audio modalities. 
        \item \textbf{Retrieval-Augmented Generation (RAG)}:
            \lclms{} simplify RAG pipelines by directly reasoning over a corpus, overcoming challenges like query decomposition~\citep{Perez2020UnsupervisedQD} and mitigating cascading errors due to retrieval misses~\citep{Barnett2024SevenFP, Longpre2021EntityBasedKC}.
        \item \textbf{SQL}:
            We explore \lclms{}' capacity to process entire databases as text, enabling natural language database querying and bypassing conversion to a formal query language like SQL~\cite{zhong2017seq2sql}.
            This potentially enables more expressive querying and handling of noisy or mixed-structured data. Importantly, it can also be seen as a case study representing the ability of \lclms{} to subsume other types of structured data and complicated formal languages to query them, such as knowledge graphs that often require bespoke solutions. 
        \item \textbf{Many-Shot ICL}:
            \lclms{} can scale the number of examples from the tens in the traditional in-context learning setup to hundreds or thousands~\citep{yu2020dialogue,Srivastava2022BeyondTI}, removing the need to find the optimal set of few-shot examples to use~\citep{Luo2023DrICLDI}.
        \end{itemize}

The \benchmark{} benchmark opens up a novel line of research on long-context prompting, which we introduce as Corpus-in-Context (CiC) Prompting (\Cref{section:prompting}). Using this approach, we evaluate Gemini 1.5 Pro~\citep{Reid2024Gemini1U}, GPT-4o~\citep{Achiam2023GPT4TR}, and Claude 3 Opus~\citep{TheC3} on \benchmark{}. Figure~\ref{fig:overview} summarizes the performance of these \lclms{} and specialized models that were carefully hand-optimized for each task, showcasing how LCLMs can tackle \benchmark{} tasks without specialized pipelines.

Our evaluation on \benchmark{} reveals several key insights when comparing state-of-the-art \lclms{} with specialized, task-specific models. At the 128k token level, the largest size comparable across all models, \lclms{} rival the performance of Gecko~\cite{Lee2024GeckoVT}, a leading textual retrieval system. Notably, Gemini~\cite{Reid2024Gemini1U} also surpasses strong multi-modal retrieval models such as CLIP~\cite{clip}.   However, \lclms{} lag significantly on complex multi-hop compositional reasoning tasks, indicating substantial room for improvement.  
Furthermore, rigorous ablations reveal large performance variance depending on prompting strategies such as chain-of-thought reasoning, underscoring the need for further research to enhance \lclms{} robustness and instructability.
Taken together, our results on \benchmark{} demonstrate that \lclms{} can match the performance of many specialized models, while also revealing ample headroom for improvement in robust long-context reasoning as context windows continue to scale.

\vspace{-5pt}
\section{\benchmark: A 1 Million+ Token Long-Context Benchmark}\label{section:dataset}
\vspace{-5pt}

\begin{table}
    \centering
    \setlength\tabcolsep{3pt}
     \setlength\extrarowheight{-1pt}
     \resizebox{0.92\linewidth}{!}{
    \begin{tabular}{cllccccc}
    \toprule
    \textbf{Task} & \textbf{Dataset} & \textbf{Description} & \textbf{\makecell{Avg.  Cand.\\Length}} & \textbf{\makecell{\# Cand. \\  (128k)}} & \textbf{\makecell{Candidates}} & \textbf{Input} & \textbf{Target} \\
    \midrule
    \textbf{\multirow{13}*{\makecell{Text\\Retrieval}}} & ArguAna & Argument Retrieval & 196 & 531 & \multirow{13}*{Passages}  & \multirow{13}*{Query} & \multirow{13}*{Passage ID(s)} \\
    & FEVER & Fact Checking & 176  & 588 & \\
    & FIQA &  Question Answering  & 196 &	531 & \\
    & MS MARCO & Web Search & 77 &	1,174 & \\
    & NQ & Question Answering & 110 & 883 & \\
    & Quora & Duplication Detection & 14 & 3,306 & \\
    & SciFact & Citation Prediction & 301 &	357 & \\
    & Touché-2020 & Argument Retrieval & 330 &	329 & \\
    & TopiOCQA & Multi-turn QA & 149 & 680 & 	\\

   & HotPotQA & Multi-hop QA &  74 &	1,222   &  \\
    & MuSiQue & Multi-hop QA & 120 &	824 \\
    & QAMPARI & Multi-target QA & 132 &	755  \\
    & QUEST & Multi-target QA & 328 &	328  \\
    \midrule

    \textbf{\multirow{4}*{\makecell{Visual \\Retrieval}}} & Flickr30k & Image Retrieval & 258 & 440 & \multirow{1}*{Images} &  \multirow{1}*{Text Query} & \multirow{1}*{Image ID} \\
    & MS COCO &  Image Retrieval & 258 & 440 & \multirow{1}*{Images} &  \multirow{1}*{Text Query} & \multirow{1}*{Image ID} \\
    & OVEN  & Image-text Retrieval & 278 & 448 & Images+Texts  & Image+Text Query  & Wikipedia ID \\
    & MSR-VTT & Video Retrieval & 774 & 140 & Videos  & Text Query & Video ID \\
    \midrule
    \textbf{\multirow{5}*{\makecell{Audio\\Retrieval}}} & FLEURS-en  & \multirow{5}*{\makecell{Audio Retrieval}} & 249 & 428 &  \multirow{5}*{\makecell{Speech}} &   \multirow{5}*{Text Query} &  \multirow{5}*{\makecell{Speech ID}} \\
    & FLEURS-es  & & 315 & 343 \\
    & FLEURS-fr  &  & 259 & 412 \\
    & FLEURS-hi  & & 292 & 369 \\
    & FLEURS-zh  & & 291 & 370 \\
    \midrule
    \textbf{\multirow{6}*{\makecell{RAG}}}  & NQ & Question Answering &  110 &	883  & \multirow{6}*{Passages} &  \multirow{6}*{Question} & \multirow{6}*{Answer(s)} \\
    & TopiOCQA & Multi-turn QA& 149 & 680\\

    & HotPotQA & Multi-hop QA & 74 & 1,222  \\
    & MuSiQue & Multi-hop QA & 120 &	824 \\
    & QAMPARI & Multi-target QA & 132	& 755  \\
    & QUEST & Multi-target QA & 328 & 328  \\
    \midrule
    \textbf{\multirow{2}*{SQL}} & Spider & Single-turn SQL & 111k & 1 & \multirow{2}*{\makecell{SQL\\Database}} &   \multirow{2}*{Question} &  \multirow{2}*{\makecell{Answer}} \\
    & SParC & Multi-turn SQL & 111k & 1  &  \\
    \midrule
    \textbf{\multirow{5}*{\makecell{Many-Shot\\ICL}}} & BBH-date  & \multirow{1}*{Multiple-choice QA}  &  131 & 150 & \multirow{5}*{\makecell{Training\\Examples}} &   \multirow{5}*{Question} &  \multirow{5}*{\makecell{Answer}}  \\
    & BBH-salient  &  Multiple-choice QA & 246 & 104 &  \\
    & BBH-tracking7 & Multiple-choice QA & 205 & 123 &  \\
    & BBH-web & Multiple-choice QA & 43 & 150 &  \\
    & LIB-dialogue &  Classification  & 266 & 274  \\
    \bottomrule
    \\
    \end{tabular}
    }
    
    \caption{
        \label{tab:dataset_statistics}
        Tasks and datasets in the \benchmark{} benchmark. \benchmark{} has 6 types of tasks, 4 modalities, and 35 datasets in total.
        For each dataset, we show the average length of the candidates (Avg. Cand. Length) as well as the number of candidates (\# Cand) in the 128k version of \benchmark{}.
        More details on the datasets are available in \Cref{appendix:dataset-selection,appendix:dataset_processing,appendix:dataset_stats_detail}.
    }
    \vspace{-20pt}
\end{table}

The \benchmark{} benchmark aims to cover a wide range of real-world applications where LCLMs can be employed. These tasks range from retrieving relevant documents for a query to extracting compositional information from databases. \Cref{tab:dataset_statistics} lists all tasks and their corresponding datasets.
    
For each dataset in all tasks, we sample up to 100 test queries, 5 few-shot queries, and  10 development queries.
To test how \lclm{}s perform when scaling the number of tokens in their context, we create \benchmark{} with three different context length limits, namely 32k\footnote{Since the gold documents of 100 test queries alone often exceed 32k tokens, we do not include test queries for the 32k version. We report the development set performance for 32k instead.}, 128k, and 1M.
\benchmark{} currently supports 1 million token contexts, but with the increasing capabilities of state-of-the-art \lclms{}, our method allows \benchmark{} to easily scale to larger context lengths (e.g., 1 billion tokens) in the future.
To allow testing the same set of queries over different context lengths, we process each dataset to have the same evaluation queries across different context lengths (except for SQL as detailed below).
    \\

\vspace{-10pt}
\vspace{-7pt}
    \paragraph{Retrieval \& RAG}
    We include diverse text retrieval and RAG datasets, covering heterogeneous retrieval tasks from BEIR~\citep{Thakur2021BEIRAH}, multi-turn conversational QA~\citep{Adlakha2021TopiOCQAOC}, multi-hop QA~\citep{Yang2018HotpotQAAD,Trivedi2021MM}, as well as multi-target QA that requires set operations such as unions or differences~\citep{Amouyal2022QAMPARIAB, Malaviya2023QUESTAR}. 
    For retrieval, we also include multimodal datasets, covering image, video, and audio~\citep{flickr,Xu2016MSRVTTAL,Conneau2022FLEURSFL}. 
    
    \begin{wrapfigure}{tr}{0.35\textwidth}
    \centering
    \includegraphics[width=0.34\columnwidth]{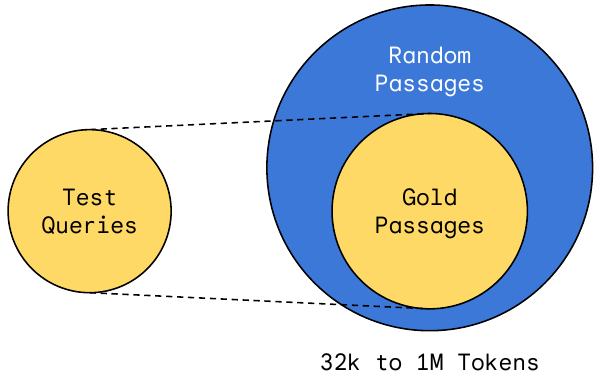}
    \caption{
        \label{fig:corpus_construction}
        \textbf{Corpus creation} for retrieval and RAG.
        Given a set of test queries, we use their associated gold passages and other random passages to form the corpus.
    }
    \vspace{-0.5cm}
\end{wrapfigure}

    All queries in each retrieval and RAG dataset share a single corpus, mimicking real retrieval applications. To create this shared corpus, we first include all gold passages from few-shot, development and the test queries, and then sample passages uniformly until reaching the desired context size (Figure~\ref{fig:corpus_construction}). This construction ensures smaller corpora (e.g., 128k) are subsets of larger ones (e.g., 1M).
    Gold and random passages are shuffled to avoid positional biases.
    For fair comparison, specialized retriever models also use the same corpora for the evaluation.
    
    \vspace{-3pt}
    \paragraph{Many-shot ICL}
    We adapt datasets from Big-Bench Hard (BBH)~\citep{Srivastava2022BeyondTI, Suzgun2022ChallengingBT} and LongICLBench (LIB)~\citep{yu2020dialogue,Li2024LongcontextLS} to evaluate many-shot in-context learning (ICL) capabilities.
    Similar to retrieval, we construct shared many-shot ICL contexts, ensuring training examples in smaller contexts are included in larger ones.
    Since all of the many-shot ICL datasets are classification tasks, we guarantee that each class is represented at least once. %
    
    \vspace{-3pt}
    \paragraph{SQL}
    We evaluate SQL-like reasoning on Spider, a single-turn text-to-SQL dataset \cite{Yu2018SpiderAL}, and SparC, its multi-turn variant \cite{Yu2019SParCCS}. 
    The corpus for each query is an associated database of one or more tables. To construct the corpus for each context length, we must select databases such that all of its tables are no larger than that context length. We do this by selecting the largest databases that will fit into that context. This necessarily means that the databases for the 1M token setting would not fit into the smaller context lengths. Therefore, unlike most of the other tasks that share a corpus, the query sets differ across \benchmark{} sizes.

    Given a maximum context length of $N \in \{\text{32k}, \text{128k}, \text{1M}\}$, we create a corpus up to a size of $0.9N$, to account for differences in tokenizers, as well as to reserve room for instructions and formatting, which will be explained in more detail in \Cref{section:prompting}.
    Please refer to \Cref{appendix:dataset-selection} for more details about how the datasets are selected for each task.

\vspace{-3pt}
\section{\Method{}}\label{section:prompting}
\vspace{-4pt}
    Traditionally, utilizing large corpora of passages, data tables, or training examples requires specialized recipes or systems. 
    \lclms{} now enable direct ingestion and processing of entire corpora within their context window. This unlocks a novel prompting-based approach for solving new and existing tasks, which we call \textbf{C}orpus-\textbf{i}n-\textbf{C}ontext prompting (\textbf{\method{}}, pronounced "seek").

    \vspace{-5pt}
    \subsection{Prompt Design}
\begin{figure}[t!]
    \centering
    \includegraphics[width=1.0\columnwidth]{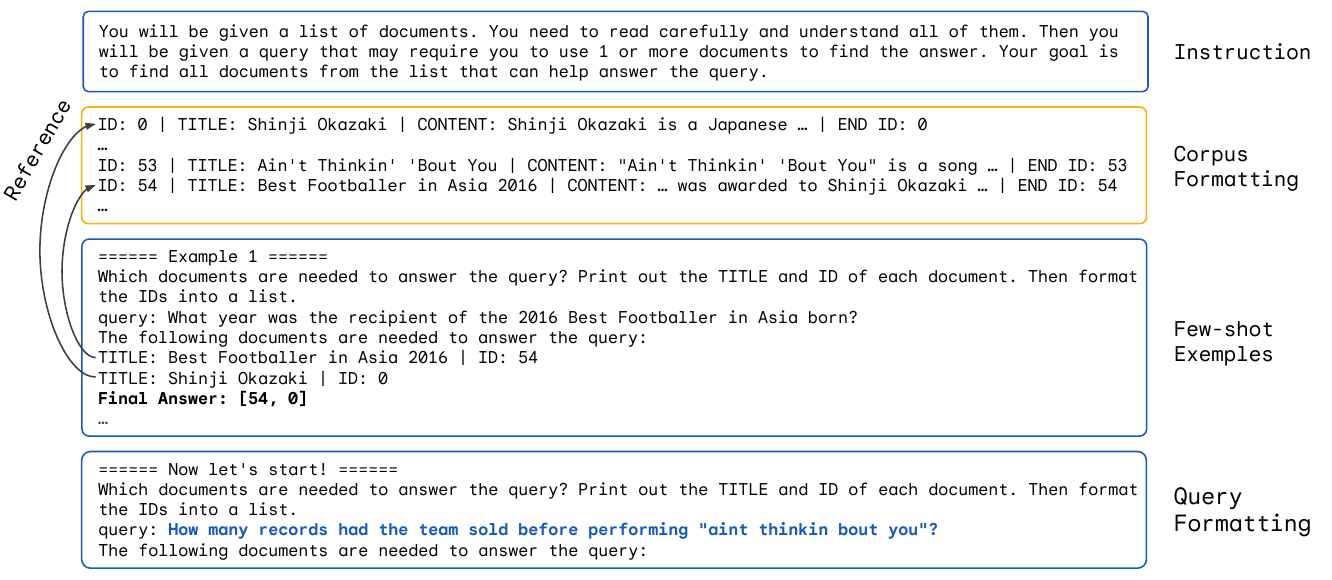}
    \vspace{-10pt}
    \caption{
        \label{fig:prompt_design}
        \textbf{Example of \Method{}} for retrieval.
        \method{} prompting leverages large language models' capacity to follow instructions, leverage few-shot examples, and benefit from reasoning demonstrations to retrieve and reason over large corpora provided in context.
    }
    \vspace{-10pt}
\end{figure}

    \vspace{-3pt}
    \method{} prompting effectively combines established prompting strategies, tailoring them to leverage the unique capabilities of \lclms{} for learning, retrieving and reasoning over in-context corpora. \Cref{fig:prompt_design} illustrates our key design choices, whose effectiveness is rigorously evaluated through extensive ablation studies in \Cref{section:more_ablations}.
    
    \vspace{-7pt}
    \paragraph{Instructions} We first provide task-specific instructions to guide the \lclm{}'s behaviors~\citep{Kojima2022LargeLM, Wei2021FinetunedLM, Chung2022ScalingIL}.
    As an example, for the retrieval task shown in \Cref{fig:prompt_design}, we ask the model to read the corpus carefully and find relevant documents to answer the question.
    \vspace{-7pt}
    \paragraph{Corpus Formatting} We then insert the entire corpus into the prompt.
    Each candidate (e.g., passage, image, audio) in a corpus is assigned a unique identifier (ID) that can be referenced as needed for that task.\footnote{Original candidate IDs from each dataset were not used as they inadvertently disclose the gold documents.} For instance, in retrieval tasks, the \lclms{} must output the correct candidate IDs.
    The structure of the corpus significantly impacts retrieval performance.
    Careful formatting, such as putting document IDs both before and after the passage in text retrieval, can mitigate the effects of causal attention in decoder-only \lclms{} and enhance retrieval accuracy.
    \vspace{-7pt}
    \paragraph{Few-Shot Examples} Providing a limited number of demonstrations helps the \lclm{} grasp the desired response format and improves task accuracy~\citep{Brown2020LanguageMA}. Unlike common approaches where few-shot examples are independent,  we ground all examples to the same corpus, aiming to teach the model to also learn more details about the specific corpus it needs to use.\footnote{As we will see in \Cref{section:results}, carefully positioning the answers of these few-shot examples can guide the model's attention to areas where it is typically weaker, mitigating "dead zones" in attention distribution.}
    To facilitate automated evaluation, answers within each few-shot example are formatted as a list (e.g., "Final Answer: [54, 0]" in \Cref{fig:prompt_design}), thus guiding the model to generate responses in a similar structure that can be readily parsed and compared against ground truth labels.
    Finally, each few-shot example is accompanied by a Chain-of-Thought reasoning ~\citep{Nye2021ShowYW, Wei2022ChainOT}.
    We find adding Chain-of-Thought reasoning helps the most on tasks requiring complex multi-hop compositional reasoning.
    
    \paragraph{Query Formatting} The query to be evaluated is formatted similar to the few-shot examples. For multi-turn datasets, we prepend previous query turns and model outputs to the current query turn, ensuring that the model's generation is conditioned on its prior responses.
    Based on our query formatting, \lclms{} generate tokens that are parsed into the final answer.

        \Cref{appendix:dataset_instructions} provides examples of instructions for all datasets in \benchmark{}, along with details on the chain-of-thought reasonings used in this paper.

\vspace{-5pt}
\subsection{Design Consideration}
    Given the variation in instructions, format, and tokenizers across different \method{} prompting techniques, the resulting context lengths can differ substantially.
    To accommodate this diversity, we allocate ample space for prompt customization, as detailed in \Cref{section:dataset}.
    To ensure a fair comparison among \lclms{}, we strongly recommend that for each maximum context length of \benchmark{} (e.g., 32k or 128k), the model still uses only the corpus and examples present in that version. We also recommend to evaluate models on the maximum size that can fit into their context length without truncating the corpus or any of the individual examples.

\vspace{-5pt}
\subsection{Discussion on Efficiency}
\label{section:efficiency}
    Encoding a one million token context can be slow and computationally expensive.
    One key advantage of CiC prompting is its compatibility with prefix-caching in autoregressive language models~\citep{google2024caching} as the query appears at the end of the prompt.
    This means \emph{the corpus only needs to be encoded once}, similar to the indexing process in traditional information retrieval.
    As demonstrated in \Cref{section:more_ablations}, encoding the corpus as the prefix in this way does not lead to a performance drop in \lclms{}.

\begin{table}[t]
    \centering
    \small
    \setlength\extrarowheight{-1pt}
    \resizebox{0.84\linewidth}{!}
    {%
    \begin{tabular}{clccc|c}
    \toprule
    & \textbf{Dataset} & \textbf{Gemini 1.5 Pro} & \textbf{GPT-4o} & \textbf{Claude 3 Opus} & {\bf Specialized}\\
    \midrule
    \textbf{\multirow{14}*{\makecell{Text\\Retrieval}}} & ArguAna & 0.84 & 0.85  & 0.74 & 0.75 \\
    & FEVER & 0.98 & 0.96 & 0.94 & 0.97 \\
    & FIQA & 0.79 & 0.82 & 0.61 & 0.83 \\
    & MS MARCO & 0.95 & 0.87 & 0.93 & 0.97 \\
    & NQ & 1.00 & 0.99 & 0.96 & 0.99 \\
    & Quora & 0.93  & 0.93 & 0.94 & 1.00 \\
    & SciFact & 0.88 & 0.88 & 0.73 & 0.85 \\
    & Touché-2020 & 0.91 & 0.88 & 0.71 & 0.88 \\
    & TopiOCQA & 0.49 & 0.30 & 0.42 & 0.36 \\
    & HotPotQA$^\dagger$  &0.90 & 0.82 & 0.83 & 0.92 \\
    & MuSiQue$^\dagger$ & 0.42 & 0.10 & 0.27 & 0.29 \\
    & QAMPARI$^\dagger$ & 0.61 & 0.18 & 0.20 & 0.57 \\
    & QUEST$^\dagger$ & 0.30 & 0.19 & 0.18 & 0.54 \\
    & \textbf{Average} & \textbf{0.77} & 0.67 & 0.65 & 0.76 \\
    \midrule
    \textbf{\multirow{5}*{\makecell{Visual\\Retrieval}}} & Flickr30k & 0.84 & 0.65 & - & 0.75 \\
    & MS COCO & 0.77 & 0.44 & - & 0.66 \\
    & MSR-VTT  & 0.76 & 0.72 & - & 0.64 \\
    & OVEN  & 0.93 & 0.89 & - & 0.79 \\
    & \textbf{Average} & \textbf{0.83} & 0.68 & - & 0.71 \\
    \midrule
    \textbf{\multirow{6}*{\makecell{Audio\\Retrieval}}} & FLEURS-en & 1.00 & -  & - & 0.98 \\
    & FLEURS-es  & 0.99 & - & - &  0.99 \\
    & FLEURS-fr  & 1.00 & - & - &  1.00 \\
    & FLEURS-hi  & 1.00 & - & - &  0.74 \\
    & FLEURS-zh  & 1.00 & - & - &  1.00 \\
    & \textbf{Average} & \textbf{1.00} &  - & - & 0.94  \\
    \midrule
    \textbf{\multirow{7}*{\makecell{RAG}}} & NQ & 0.84 & 0.89 & 0.85 & 0.71 \\
    & TopiOCQA & 0.34 & 0.33 & 0.37 & 0.35 \\
    & HotPotQA & 0.75 & 0.72  & 0.74 & 0.70 \\
    & MuSiQue & 0.55 & 0.47 & 0.45 & 0.45 \\
    & QAMPARI & 0.44 & 0.27 & 0.25 & 0.55 \\
    & QUEST & 0.28 & 0.20 & 0.15 & 0.35 \\
    & \textbf{Average} & \textbf{0.53} & 0.48 & 0.47 & 0.52 \\
    \midrule
    \textbf{\multirow{3}*{SQL}} & Spider & 0.40 & 0.14 & 0.19 & 0.74 \\
    & SParC & 0.36 & 0.13 & 0.21 & 0.55 \\
    & \textbf{Average} & 0.38 & 0.13 & 0.20 & \textbf{0.65} \\
    \midrule
    \textbf{\multirow{6}*{\makecell{Many-Shot\\ICL}}} & BBH-date  & 0.88 & 0.81 & 0.92 & - \\
    & BBH-salient & 0.78 & 0.64 & 0.69 & - \\
    & BBH-tracking7 & 0.33 & 0.81 & 0.54 & - \\
    & BBH-web & 0.67 & 0.57 & 0.83 & - \\
    & LIB-dialogue & 0.76 & 0.67 & 0.72 & - \\
    & \textbf{Average} & 0.68 & 0.70 & \textbf{0.74} & - \\
    \bottomrule
    \\
    \end{tabular}}
    \caption{
        \label{tab:main_results}
        \textbf{Main Results on \benchmark{} 128k context test set}.
        We show performances of three LCLMs as well as specialized models that rely on task-specific fine-tuning or pipelining.
        For the evaluation metrics: text, visual, and audio retrieval use Recall@1; RAG uses subspan exact match; SQL uses accuracy; and many-shot ICL uses classification accuracy.
        $^\dagger$: For retrieval with multiple gold targets, MRecall@$k$ ($k=2, 5, 5, 3$ in order) is employed as described in \Cref{appendix:dataset-selection}.
    }
\vspace{-20pt}
\end{table}

\vspace{-8pt}
\section{\benchmark{} Tasks and Primary Results}\label{section:results}
\vspace{-5pt}
    We evaluate three state-of-the-art \lclms{} on \benchmark{}: Google's \textbf{Gemini 1.5 Pro}~\citep{Reid2024Gemini1U}, OpenAI's \textbf{GPT-4o}~\citep{Achiam2023GPT4TR}, and Anthropic's \textbf{Claude 3 Opus}~\citep{TheC3}.
    We evaluate these models because their APIs support most of the modalities.
    Their maximum context lengths are 2M, 128k, and 200k tokens, respectively. 
    We use their official APIs\footnote{\url{https://ai.google.dev/gemini-api}}$^,$\footnote{\url{https://platform.openai.com/docs/models/gpt-4o}}$^,$\footnote{\url{https://www.anthropic.com/api}} for the evaluation.
    Their prompts were chosen based on their performance on the development queries over the 128k token context.
    While evaluating, a small number of API calls can be blocked due to various reasons such as errors in the safety filters; these examples are treated as incorrect.
    In all \benchmark{} tasks, \textbf{\lclms{} without any task-specific fine-tuning are benchmarked against specialized models that have undergone extensive fine-tuning or pipelining for the target task} and are therefore limited to that specific domain.
    We select each specialized model that exemplifies recent task-specific advancements.

\subsection{Text Retrieval}
\vspace{-5pt}

    \begin{wrapfigure}{tl}{0.35\textwidth}
    \centering
    \vspace{-13pt}
    \includegraphics[width=0.35\columnwidth]{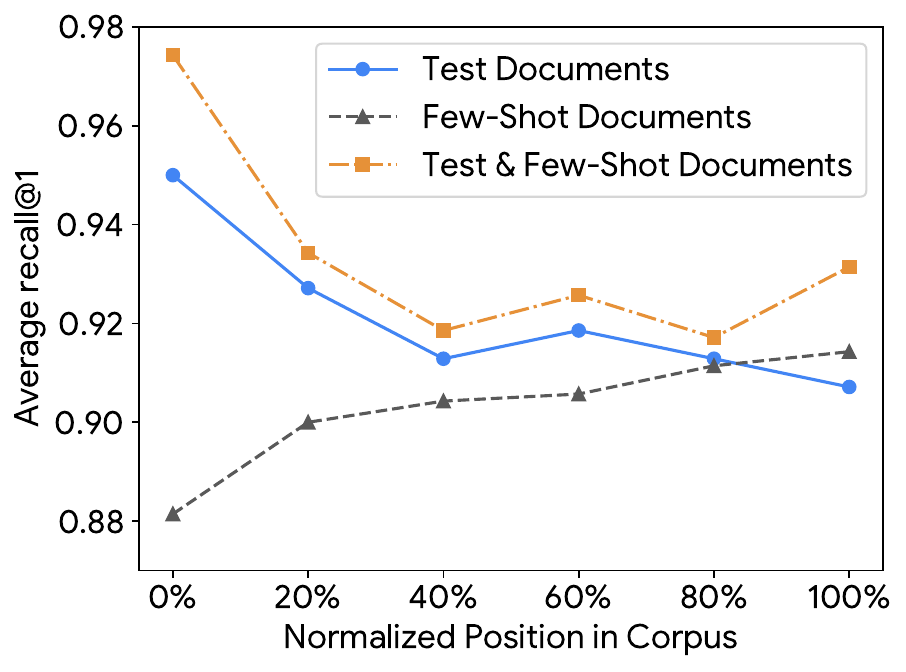}
    \vspace{-17pt}
    \caption{
        \label{fig:position-experiments-average}
        \textbf{Positional Analysis}.
        We vary gold document positions of queries within the corpus (0\% = beginning, 100\% = end). 
    }
    \vspace{-20pt}
\end{wrapfigure}

\begin{figure}[t]
    \centering
    \includegraphics[width=1.0\columnwidth]{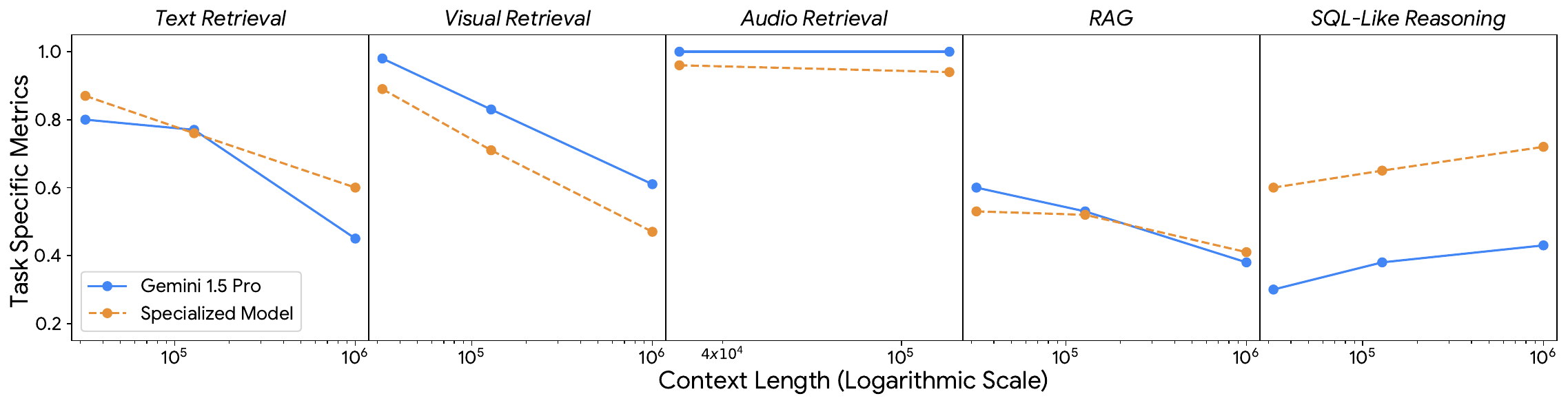}
    \vspace{-15pt}
    \caption{
        \label{fig:scaling_results}
        \textbf{Scaling results} of \lclms{} compared to each specialized model by scaling the corpus size from 32k to 1 million tokens.
        Results are averaged over all constituent datasets in each task.
    }
\vspace{-12pt}
\end{figure}

    We adopt Gecko~\citep{Lee2024GeckoVT}, a state-of-the-art dual encoder as the specialized model for the retrieval task.
    Gecko is fine-tuned on extensive text retrieval and similarity tasks.
    To ensure fair comparison, we use exactly the same corpus to test both the \lclms{} and Gecko, instead of using the full retrieval corpora or published results.
    \paragraph{Results} \Cref{tab:main_results} demonstrates that Gemini 1.5 Pro performs comparably to Gecko at 128k context length.
    Other \lclms{} also perform surprisingly well.
    This is notable, as \lclms{} have not undergone specialized contrastive learning for retrieval. While \lclms{}'s performance does degrade when scaling the corpus to millions of tokens (\Cref{fig:scaling_results}), the parity at 128k still suggests the potential of \lclms{} to be used for retrieval tasks.

    \paragraph{Positional Analysis} 
    To better understand the performance of \lclms{} on longer context lengths, we investigate how the positioning of the gold document affects retrieval performance, examining the effect of the gold document for the test query and for the few shot examples within the prompt~\cite{Liu2023LostIT}.

    \Cref{fig:position-experiments-average} reveals that performance drops as the gold documents of the test queries are moved towards the end of the corpus, suggesting reduced attention in later sections of the prompt.
    Conversely, placing the gold documents of few-shot queries at the end improves recall, indicating their ability to mitigate attention weaknesses in this region.
    Co-locating gold documents of few-shot and test queries consistently boosts performance. This is perhaps unsurprising, as it gives the model information about where to look for the answer; however, it also indicates that the model does indeed pay special attention to the locations where the gold documents for the few-shot examples are placed, regardless of where they are in the corpus. This offers a promising approach to overcome performance degradation in large corpora.
    Per-dataset analysis is provided in \Cref{appendix:positional-analysis}.

\vspace{-5pt}
\subsection{Visual Retrieval}

    \vspace{-5pt}
    We employ CLIP-L/14~\citep{clip}, a widely used text-to-image retrieval model, as our specialized model. For Flickr30k and MS-COCO, CLIP performs text-to-image retrieval. For MSR-VTT, it performs text-to-video retrieval by averaging scores across frames. For OVEN, due to the lack of suitable open-source image-to-text models, we approximate image-to-text retrieval by using CLIP's text-to-image retrieval.
    Evaluation of Claude 3 Opus on this task was not feasible due to the current limitation of 20 images per API request.

    \vspace{-5pt}
    \paragraph{Results} Gemini 1.5 Pro outperforms GPT-4o across all four visual benchmarks (\Cref{tab:main_results}).
    Notably, as shown in Figure~\ref{fig:scaling_results}, Gemini 1.5 Pro maintains a performance advantage over CLIP across all visual benchmarks and context lengths.

\vspace{-5pt}
\subsection{Audio Retrieval}
    
    \vspace{-5pt}
    We choose PaLM 2 DE~\citep{gomez2024transforming} as a specialized model, which is a dual-encoder trained to maximize the similarity between audio and their transcription and has achieved previous state-of-the-art on the FLEURS datasets.
    Currently, GPT-4o and Claude 3 Opus do not support audio input.

    \vspace{-5pt}
    \paragraph{Results}  Gemini 1.5 Pro demonstrates comparable performance to PaLM 2 DE across all 5 languages (\Cref{tab:main_results}). We notice that Gemini 1.5 Pro notably surpasses PaLM 2 DE in Hindi; this advantage likely stems from differences in pre-training data between Gemini and PaLM.   \Cref{fig:scaling_results} further confirms Gemini 1.5 Pro's robust performance across various context length, highlighting the current capabilities of \lclms{} while also indicating the need for more challenging audio datasets.

\vspace{-5pt}
\subsection{RAG}
\vspace{-5pt}
    We set up a retrieve-and-read RAG pipeline as a specialized model, using Gecko~\cite{Lee2024GeckoVT} to retrieve the top-40 documents which are then put into the context of Gemini 1.5 Pro and used to generate the answer conditioned on the question and the retrieved documents.

    \vspace{-5pt}
    \paragraph{Results}
    \Cref{tab:main_results} demonstrates that Gemini 1.5 Pro, with the entire corpus in context, outperforms the RAG pipeline on multi-hop datasets (HotpotQA and MusiQue). This is because \lclms{} can reason over multiple passages in the context window using Chain-of-Thought~\cite{Wei2022ChainOT}, a capability that RAG pipelines typically lack unless they have a separate module for planning and reasoning.

    However, a specialized retriever like Gecko excels at ranking all topically relevant passages from a corpus, enabling it to identify a comprehensive set of passages covering all answers. This proves particularly beneficial for multi-target datasets, such as QUEST and QAMPARI.

    \Cref{fig:scaling_results} reveals that while \lclms{} match RAG performance at 128k compared to a pipeline, performance drops at 1M corresponding to the drop found in \lclm{} text retrieval performance.

    \begin{wraptable}{r}{0.40\linewidth}
\small
\vspace{-12pt}
\begin{tabular}{ccc}
\toprule  
\textbf{Dataset} & \textbf{Dev (32k)} & \textbf{Test (128k)}\\
\midrule
NQ & 0.60 \textcolor{red}{(-0.10)} & 0.37 \textcolor{red}{(-0.47)} \\  
HotPotQA & 0.60 \textcolor{red}{(-0.30)} & 0.33 \textcolor{red}{(-0.42)} \\
MuSiQue & 0.20 \textcolor{red}{(-0.60)} & 0.10 \textcolor{red}{(-0.45)} \\  
\bottomrule
\end{tabular}
\caption{\textbf{Gemini's closed-book performance on RAG}.
Red indicates the performance difference compared to the \method{} prompting.}
\vspace{-10pt}
\label{table:closed_book_rag}
\end{wraptable} 

    \paragraph{Closed-Book Ablations}
    To further probe \lclms{} capabilities, we conduct closed-book ablations on Gemini 1.5 Pro. In this setting, we remove the corpus from the context to assess \lclm{} performance based solely on parametric knowledge~\citep{Lewis2020QuestionAA, Longpre2021EntityBasedKC}.
    \Cref{table:closed_book_rag} presents the results, revealing that the closed-book performance significantly lags behind the long-context and specialized models.
    This underscores the tested models' effectiveness in leveraging the external corpus to enhance its reasoning capabilities.

\subsection{SQL-Like Compositional Reasoning}
\vspace{-0pt}

    \begin{wrapfigure}{tl}{0.34\columnwidth}
    \centering
    \vspace{-10pt}
    \includegraphics[width=0.34\columnwidth]{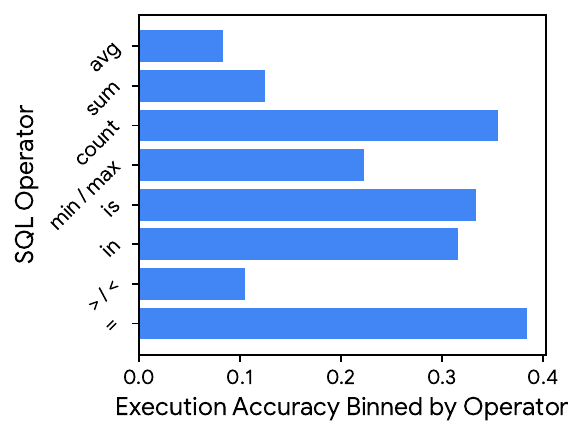}
    \caption{
        \textbf{SQL Reasoning Analysis}.
        We bin Spider queries by operators in their SQL query and report binned Gemini performance.
        We group \texttt{min} and \texttt{max} into a bin and  \texttt{>} and \texttt{<} into another bin.
    }
    \vspace{-15pt}
    \label{fig:sql_reasoning_analysis}
\end{wrapfigure}

    The traditional SQL pipeline uses a trained semantic parser to translate the natural language input into a SQL query.
    Then, a separate SQL interpreter is used to execute the SQL query over the database.
    As a specialized model, we use DAIL-SQL~\citep{Gao2023TexttoSQLEB} for the semantic parser, which prompts an LLM to provide the SQL query.
    We adapt DAIL-SQL by replacing its LLM with Gemini 1.5 Pro and using a fixed set of few-shot examples.
    
    \paragraph{Results}
    Results in \Cref{tab:main_results} show that \lclms{} achieve reasonable performance, though they are significantly behind the specialized pipeline. This reveals substantial headroom to enhance the compositional reasoning capabilities of \lclms{}.
    
    Lower performance than the specialized pipeline on the SQL task is unsurprising, given the decades of work on semantic parsing and SQL.
    Nonetheless, this performance still reveals the potential for the model to handle complex structured data with no task-specific tuning.
    
    \vspace{-0pt}
    \paragraph{Reasoning Analysis}
    To gain insights into the short-comings of \lclms{} in complex compositional reasoning, %
    we categorize queries based on the operators in the gold SQL queries and measure Gemini 1.5 Pro's performance for each operator.
    \Cref{fig:sql_reasoning_analysis} shows that averaging is the most difficult operation while counting is relatively easy.
    Moreover, we find that reasoning over equality is considerably easier than reasoning over inequality.
\vspace{-0pt}
\subsection{Many-Shot ICL}
    \begin{wrapfigure}{tl}{0.34\columnwidth}
    \centering
    \vspace{-15pt}
    \includegraphics[width=0.34\columnwidth]{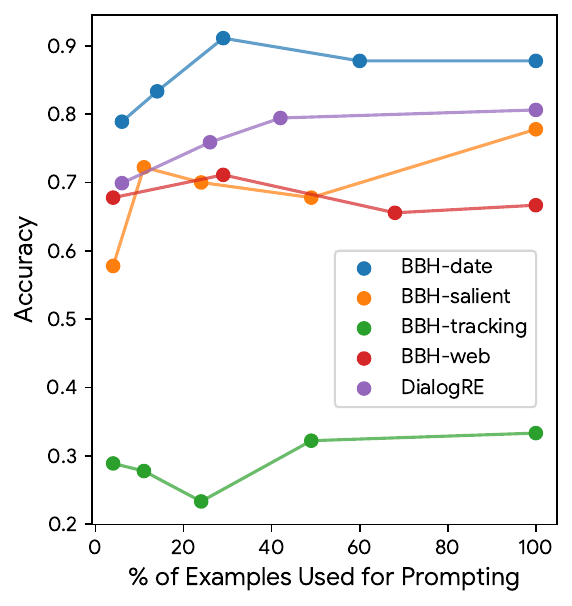}
    \vspace{-10pt}
    \caption{
        \label{fig:many_shot_scaling}
        \textbf{ICL Performance} as we scale the percentage of examples used up to 100\%.
    }
    \vspace{-10pt}
\end{wrapfigure}

    \paragraph{Results}
    \cref{tab:main_results} compares accuracy for each of the \lclms{} across all ICL datasets.
    For BBH, we report the accuracy on 32k, which is the maximum context length available.
    Gemini 1.5 Pro outperforms GPT-4o on all benchmarks, except for BBH-tracking7 where Gemini performs surprisingly poorly.
    On average, Claude 3 Opus achieves the best performance among \lclms{} on this task.
    
    \paragraph{Scaling Many Shot ICL}
    \Cref{fig:many_shot_scaling} illustrates the impact of increasing the number of examples in Gemini.
    In LIB-dialog, accuracy improves monotonically with more examples.
    In contrast, results on BBH are mixed.
    Knowledge-intensive tasks like BBH-date and BBH-salient see monotonic improvements similar to LIB-dialog, while reasoning-intensive tasks like BBH-tracking7 and BBH-web do not benefit.
    These results suggest that more complicated tasks may see an earlier limit in how much models can learn from scaling the number of in-context examples.

\vspace{-5pt}
\section{CiC Prompt Ablations}\label{section:more_ablations}
\vspace{-5pt}

We conduct ablations over the different facets of the CiC Prompt. Examples of the prompt for each ablation can be found in \Cref{appendix:ablated-prompt-examples}.
For the ablations, we evaluate Gemini 1.5 Pro on the 128k version of \benchmark{}.

\begin{table}
    \centering
    \resizebox{1.0\linewidth}{!}{%
    \begin{tabular}{cc|c|ccccccc}
    \toprule
    \thead{\bf Task\\ \bf(Metric)} & \textbf{Dataset} & \thead{\bf Best \\ \bf Prompt} & \thead{\bf Generic \\ \bf Instruction} & \thead{\bf Query at \\ \bf Beginning} & \thead{\bf Alphanu- \\ \bf meric IDs}  & \thead{\bf Titles \\ \bf Only}  & \thead{\bf Without \\ \bf ID Echo} & \thead{\bf Corpus in \\ \bf Each Few-shot} & \thead{\bf Without \\ \bf CoT} \\
    \midrule
    \textbf{\multirow{7}*{\makecell{Text\\Retrieval}}} & ArguAna & 0.84 & 0.76 & 0.72 & 0.81 & - & 0.78 & 0.62 & 0.79 \\
    & FIQA & 0.79 & 0.77 & 0.58 & 0.75 & - & 0.76 & 0.78 & 0.85 \\
    & NQ & 1.00 & 0.98 & 0.98 & 0.99 & 0.91 & 1.00 & 1.00 & 1.00 \\
    & SciFact & 0.88 & 0.88 & 0.81 & 0.90 & 0.84 & 0.87 & 0.78 & 0.90 \\
    & MuSiQue & 0.49 & 0.44 & 0.19 & 0.44 & 0.10 & 0.36 & 0.35 & 0.43 \\
    & QAMPARI & 0.61 & 0.61 & 0.49 & 0.54 & 0.09 & 0.49 & 0.35 & 0.43 \\
    & QUEST & 0.28 & 0.28 & 0.22 & 0.30 & 0.05 & 0.27 & 0.22 & 0.30 \\
    \midrule
    \textbf{\multirow{4}*{\makecell{RAG}}}& MuSiQue & 0.55 & 0.57 & 0.39 & 0.52 & 0.23 & 0.55 & 0.78 & 0.50 \\
    & NQ & 0.84 & 0.81 & 0.76 & 0.80 & 0.39 & 0.83 & 0.83 & 0.82 \\
    & QAMPARI & 0.44 & 0.42 & 0.35 & 0.42 & 0.09 & 0.36 & 0.36 & 0.33 \\
    & QUEST & 0.28 & 0.33 & 0.17 & 0.28 & 0.02 & 0.25 & 0.27 & 0.32 \\
    \midrule
    & \textbf{Average} & 0.64 & 0.62 & 0.51 & 0.61 & 0.30 & 0.59 & 0.55 & 0.61 \\
    & \textbf{\textcolor{red}{($\Delta$)}} & - &  \textcolor{red}{(-0.02)}  & \textcolor{red}{(-0.13)} & \textcolor{red}{(-0.03)} & \textcolor{red}{(-0.30)} & \textcolor{red}{(-0.05)} & \textcolor{red}{(-0.09)} &\textcolor{red}{(-0.03)} \\
    \bottomrule \\
    \end{tabular}}
    \label{tab:prompt_ablations}
    \caption{Ablation results of Gemini 1.5 Pro on different tasks in \benchmark{} at 128k context length.
    Starting from our best prompt format (used in the rest of the experiments), individual facets of the corpus, query, and instruction are ablated to surface their relative effect on quality.
    Since ArguAna and FIQA do not have any title for each passage, we report the average and its difference without these two datasets for the \texttt{Titles Only} ablation.
    } 
    \vspace{-10pt}
\end{table}

The ablations show the effectiveness of our CiC prompting design.
Removing task-specific instructions (\texttt{Generic Instruction}) or Chain-of-Thought reasoning (\texttt{Without CoT}) both lead to worse performance. %
We also observe a performance decrease for \texttt{Corpus in Each Few-Shot}. In this setting, instead of using a shared corpus, each few-shot example has its own small corpus consisting of nine random passages and one gold passage. This performance degradation could be either because the few-shot examples help the model attend to the test corpus or because the few-shot task becomes much easier than the evaluation task.

Placing the query at the beginning of the prompt instead of at the end (\texttt{Query at Beginning}) led to a significant and consistent performance decrease across tasks and models.
This result suggests that prefix-caching actually works better than encoding the corpus conditioned on each query, which would be much more expensive.

\begin{wrapfigure}{tl}{0.34\textwidth}
    \vspace{-0pt}
    \centering
    \includegraphics[width=0.34\columnwidth]{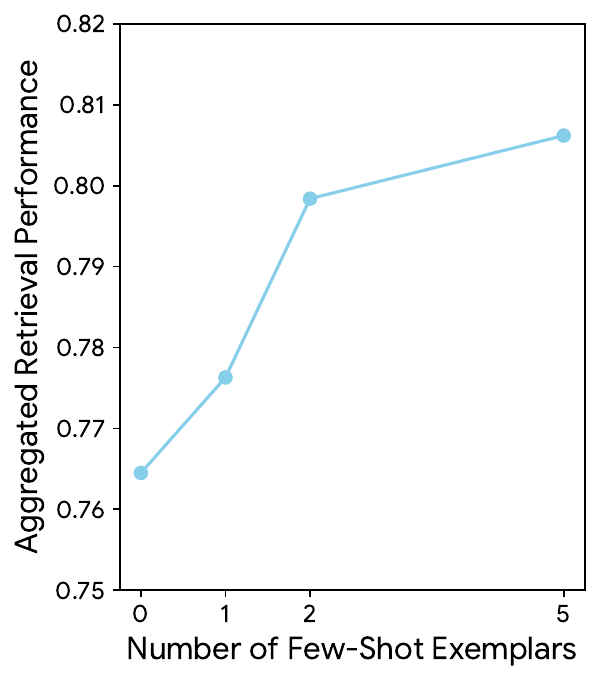}
    \vspace{-15pt}
    \caption{
        \label{fig:few-shot}
        \textbf{Effect of the number of few-shot examples}.
        The performance increases with the number of few-shot examples.
    }
    \vspace{-10pt}
\end{wrapfigure}

The presentation of document IDs also affects performance. In particular, replacing monotonic numerical IDs with random (\texttt{Alphanumeric IDs}) negatively impacts performance in most datasets. This could possibly be due to way in which numbers are tokenized, with fewer tokens for certain numbers.
Only placing the IDs at the front of the document instead of at the front and the back (\texttt{Without ID Echo}) also resulted in a 5\% performance drop, confirming that repeating text can compensate for missing context in autoregressive language models~ \citep{springer2024repetition}.

To test if the model is simply using parametric knowledge instead of grounding on the context, we remove the document content and simply keep the document title and ID in the corpus (\texttt{Title Only}). In this case, it would imply the model is able to perform well because it has already seen all of the datasets we are evaluating on during training. Across all experiments, this ablation significantly degraded performance, indicating the model indeed relies on the provided context.

We also study how the number of few-shot examples in the prompt affects quality.
The results can be found in \Cref{fig:few-shot}. 
Increasing the number of examples improves the quality on the retrieval task, from 0.76 at zero-shot to 0.81 at 5-shots.
Finally, qualitative analysis on the model outputs is provided in \Cref{appendix:qualitative}.

\vspace{-5pt}
\section{Related Work}\label{section:related_work}
\vspace{-5pt}
    
    Evaluating \lclms{} remains a challenge due to the limitations of existing benchmarks.
    Many popular datasets and methods rely on synthetic tasks \citep{Tay2020LongRA} such as the popular "Needle-in-A-Haystack" retrieval \citep{KamradtNeedle} or its extension to multi-hop QA \citep{Levy2024SameTM}.
    While these evaluations are also scalable to arbitrarily long contexts, they do not fully capture the nuances of real-world retrieval or reasoning tasks~\cite{Hsieh2024RULERWT}.
    Some recent benchmarks leverage existing NLP datasets for tasks such as extreme summarization and multi-document QA \citep{Bai2023LongBenchAB}.
    However, these tasks lack the dynamic scaling capabilities of synthetic benchmarks, which makes them difficult to adapt to very long contexts.
    
    LongAlpaca \citep{chen2024longlora} and LongBench-Chat \citep{bai2024longalign} evaluate instruction-following under long context settings but contain relatively low task diversity and no examples beyond 100k context length. Similar to \benchmark{}, Ada-LEval \citep{wang2024adaleval} proposes a length-adaptable benchmark; however, their tasks are somewhat synthetic and may not resemble real-world applications.
    
    Also related to our work is \citep{Liu2023LostIT}, which applies LCLMs to long-context QA using the top retrieved documents from Natural Questions, similar to specialized models for RAG in \benchmark{}. 
    They find that \lclms{} lose recall when relevant information is placed in the middle of the context.
    However, their analysis is limited to contexts that are under 10k tokens.
    We extend this type of evaluation of \lclms{} to context lengths of up to 1M tokens in addition to using multiple modalities and additional forms of reasoning. Finally, our work connects to the field of generative retrieval ~\cite{de2020autoregressive, tay2022transformer}, where models are trained to memorize and generate retrieval targets. Our research offers an alternative approach where the retrieval corpus is directly provided as context, eliminating task specific training.

\vspace{-10pt}
\section{Conclusion}
\vspace{-5pt}
    As language models improve and scale, their ability to retrieve and reason over increasingly long context will unlock unprecedented use-cases.
    To measure this progress, we introduce \benchmark{}, the Long Context Frontiers benchmark.
    \benchmark{} is a suite of tasks that rigorously assesses \lclms{} on tasks ripe for a paradigm shift: retrieval, retrieval-augmented generation, SQL-like reasoning, and in-context learning.
    \benchmark{} provides dynamic scaling of context lengths of up to 1 million tokens, ensuring that evaluations remain relevant as LCLMs continue to evolve. The tasks can be further scaled up with minimal effort to 1 billion tokens as context lengths continue to increase.
    Initial findings demonstrate that despite having never been trained to do retrieval, \lclms{} have retrieval capabilities rivaling task-specific hand-crafted SOTA retrieval systems.
    Nevertheless, there remains considerable room for advancement in long-context reasoning, particularly as models gain access to even longer context windows.
    We believe that \benchmark{} provides a fertile testing ground for measuring progress in long-context modeling.
    
\paragraph{Limitations} Our experiments were constrained by the computational resources and financial costs associated with utilizing \lclms{}.
The entire LOFT 128k test sets contain around $35 \text{ datasets} \times 100 \text{ prompts} \times 128\text{k} \text{ tokens}=448\text{M}$ input tokens, which cost $\$1,568$ for Gemini 1.5 Pro, $\$2,240$ for GPT-4o, and $\$6,720$ for Claude 3 Opus at the time of writing. To reduce costs, we also release dev sets, which are 10x smaller and can be evaluated with around $\$200$ using Gemini 1.5 Pro or GPT-4o.  We also expect LLM API prices to decrease over time. 
Another limitation of this work is that we focused on evaluating the quality of \lclms{}, and leave efficiency considerations for future work. We could not measure the efficiency improvements from prefix caching \citep{google2024caching} due to API constraints at the time of writing. Without caching, the Gemini 1.5 Pro API has a median latency of roughly four seconds for 32k input tokens, twelve seconds for 128k input tokens, and 100 seconds for 1 million input tokens. This speed is likely slower than specialized retrievers or SQL databases; the promising quality results on \benchmark{} encourage further investigation into optimizing \lclms{} efficiency.
Additionally,  our retrieval and RAG tasks was limited to 1 million tokens, which still leaves a large gap from real-world applications that may involve several million or even billions of documents.

\section*{Acknowledgement}
We are grateful to Frank Palma Gomez for  providing the audio retrieval model, and to Raoul de Liedekerke and Miteyan Patel for their invaluable help with the API-based model evaluation. We thank Nicholas Monath, Raphael Hoffmann, Slav Petrov, Urvashi Khandelwal, and Kristina Toutanova for their thoughtful discussions and insightful feedback to the paper.

\bibliographystyle{plain}
\bibliography{neurips_2024}

\begin{thebibliography}{10}

\bibitem{10.1162/tacl_a_00667}
Vaibhav Adlakha, Parishad BehnamGhader, Xing~Han Lu, Nicholas Meade, and Siva
  Reddy.
\newblock {Evaluating Correctness and Faithfulness of Instruction-Following
  Models for Question Answering}.
\newblock {\em Transactions of the Association for Computational Linguistics},
  12:681--699, 2024.

\bibitem{Adlakha2021TopiOCQAOC}
Vaibhav Adlakha, Shehzaad Dhuliawala, Kaheer Suleman, Harm de~Vries, and Siva
  Reddy.
\newblock {T}opi{OCQA}: Open-domain conversational question answering with
  topic switching.
\newblock {\em Transactions of the Association for Computational Linguistics},
  10:468--483, 2022.

\bibitem{Amouyal2022QAMPARIAB}
Samuel~Joseph Amouyal, Tomer Wolfson, Ohad Rubin, Ori Yoran, Jonathan Herzig,
  and Jonathan Berant.
\newblock Qampari: A benchmark for open-domain questions with many answers.
\newblock In {\em IEEE Games Entertainment Media Conference}, 2022.

\bibitem{TheC3}
Anthropic.
\newblock The claude 3 model family: Opus, sonnet, haiku.
\newblock {\em Claude-3 Model Card}, 2024.

\bibitem{bai2024longalign}
Yushi Bai, Xin Lv, Jiajie Zhang, Yuze He, Ji~Qi, Lei Hou, Jie Tang, Yuxiao
  Dong, and Juanzi Li.
\newblock Longalign: A recipe for long context alignment of large language
  models, 2024.

\bibitem{Bai2023LongBenchAB}
Yushi Bai, Xin Lv, Jiajie Zhang, Hong Lyu, Jiankai Tang, Zhidian Huang,
  Zhengxiao Du, Xiao Liu, Aohan Zeng, Lei Hou, Yuxiao Dong, Jie Tang, and
  Juanzi Li.
\newblock Longbench: A bilingual, multitask benchmark for long context
  understanding.
\newblock {\em ArXiv preprint}, abs/2308.14508, 2023.

\bibitem{Bajaj2016MSMA}
Payal Bajaj, Daniel~Fernando Campos, Nick Craswell, Li~Deng, Jianfeng Gao,
  Xiaodong Liu, Rangan Majumder, Andrew McNamara, Bhaskar Mitra, Tri~Minh
  Nguyen, Mir Rosenberg, Xia Song, Alina~Mihaela Stoica, Saurabh Tiwary, and
  Tong Wang.
\newblock Ms marco: A human generated machine reading comprehension dataset.
\newblock {\em arXiv: Computation and Language}, 2016.

\bibitem{Barnett2024SevenFP}
Scott Barnett, Stefanus Kurniawan, Srikanth Thudumu, Zach Brannelly, and
  Mohamed Abdelrazek.
\newblock Seven failure points when engineering a retrieval augmented
  generation system.
\newblock {\em ArXiv preprint}, abs/2401.05856, 2024.

\bibitem{beltagy2020longformer}
Iz~Beltagy, Matthew~E Peters, and Arman Cohan.
\newblock Longformer: The long-document transformer.
\newblock {\em ArXiv preprint}, abs/2004.05150, 2020.

\bibitem{Srivastava2022BeyondTI}
BIG bench authors.
\newblock Beyond the imitation game: Quantifying and extrapolating the
  capabilities of language models.
\newblock {\em Transactions on Machine Learning Research}, 2023.

\bibitem{Bondarenko2020OverviewOT}
Alexander Bondarenko, Maik Fr{\"o}be, Meriem Beloucif, Lukas Gienapp, Yamen
  Ajjour, Alexander Panchenko, Christian Biemann, Benno Stein, Henning
  Wachsmuth, Martin Potthast, and Matthias Hagen.
\newblock Overview of touch{\'e} 2020: Argument retrieval.
\newblock In {\em Conference and Labs of the Evaluation Forum}, 2020.

\bibitem{Brown2020LanguageMA}
Tom~B. Brown, Benjamin Mann, Nick Ryder, Melanie Subbiah, Jared Kaplan,
  Prafulla Dhariwal, Arvind Neelakantan, Pranav Shyam, Girish Sastry, Amanda
  Askell, Sandhini Agarwal, Ariel Herbert{-}Voss, Gretchen Krueger, Tom
  Henighan, Rewon Child, Aditya Ramesh, Daniel~M. Ziegler, Jeffrey Wu, Clemens
  Winter, Christopher Hesse, Mark Chen, Eric Sigler, Mateusz Litwin, Scott
  Gray, Benjamin Chess, Jack Clark, Christopher Berner, Sam McCandlish, Alec
  Radford, Ilya Sutskever, and Dario Amodei.
\newblock Language models are few-shot learners.
\newblock In Hugo Larochelle, Marc'Aurelio Ranzato, Raia Hadsell,
  Maria{-}Florina Balcan, and Hsuan{-}Tien Lin, editors, {\em Advances in
  Neural Information Processing Systems 33: Annual Conference on Neural
  Information Processing Systems 2020, NeurIPS 2020, December 6-12, 2020,
  virtual}, 2020.

\bibitem{chen2024longlora}
Yukang Chen, Shengju Qian, Haotian Tang, Xin Lai, Zhijian Liu, Song Han, and
  Jiaya Jia.
\newblock Longlora: Efficient fine-tuning of long-context large language
  models, 2024.

\bibitem{Chung2022ScalingIL}
Hyung~Won Chung, Le~Hou, Shayne Longpre, Barret Zoph, Yi~Tay, William Fedus,
  Yunxuan Li, Xuezhi Wang, Mostafa Dehghani, Siddhartha Brahma, Albert Webson,
  Shixiang~Shane Gu, Zhuyun Dai, Mirac Suzgun, Xinyun Chen, Aakanksha
  Chowdhery, Alex Castro-Ros, Marie Pellat, Kevin Robinson, Dasha Valter,
  Sharan Narang, Gaurav Mishra, Adams Yu, Vincent Zhao, Yanping Huang, Andrew
  Dai, Hongkun Yu, Slav Petrov, Ed~H. Chi, Jeff Dean, Jacob Devlin, Adam
  Roberts, Denny Zhou, Quoc~V. Le, and Jason Wei.
\newblock Scaling instruction-finetuned language models.
\newblock {\em Journal of Machine Learning Research}, 25(70):1--53, 2024.

\bibitem{Conneau2022FLEURSFL}
Alexis Conneau, Min Ma, Simran Khanuja, Yu~Zhang, Vera Axelrod, Siddharth
  Dalmia, Jason Riesa, Clara Rivera, and Ankur Bapna.
\newblock Fleurs: Few-shot learning evaluation of universal representations of
  speech.
\newblock {\em 2022 IEEE Spoken Language Technology Workshop (SLT)}, pages
  798--805, 2022.

\bibitem{de2020autoregressive}
Nicola De~Cao, Gautier Izacard, Sebastian Riedel, and Fabio Petroni.
\newblock Autoregressive entity retrieval.
\newblock {\em arXiv preprint arXiv:2010.00904}, 2020.

\bibitem{dua-etal-2019-drop}
Dheeru Dua, Yizhong Wang, Pradeep Dasigi, Gabriel Stanovsky, Sameer Singh, and
  Matt Gardner.
\newblock {DROP}: A reading comprehension benchmark requiring discrete
  reasoning over paragraphs.
\newblock In {\em Proceedings of the 2019 Conference of the North {A}merican
  Chapter of the Association for Computational Linguistics: Human Language
  Technologies, Volume 1 (Long and Short Papers)}, pages 2368--2378,
  Minneapolis, Minnesota, 2019. Association for Computational Linguistics.

\bibitem{Gao2023TexttoSQLEB}
Dawei Gao, Haibin Wang, Yaliang Li, Xiuyu Sun, Yichen Qian, Bolin Ding, and
  Jingren Zhou.
\newblock Text-to-sql empowered by large language models: A benchmark
  evaluation.
\newblock {\em ArXiv preprint}, abs/2308.15363, 2023.

\bibitem{gomez2024transforming}
Frank~Palma Gomez, Ramon Sanabria, Yun-hsuan Sung, Daniel Cer, Siddharth
  Dalmia, and Gustavo~Hernandez Abrego.
\newblock Transforming llms into cross-modal and cross-lingual
  retrievalsystems.
\newblock In {\em Proceedings of the 21st International Conference on Spoken
  Language Translation (IWSLT 2024)}, 2024.

\bibitem{google2024caching}
Google.
\newblock Context caching guide.
\newblock \url{https://ai.google.dev/gemini-api/docs/caching}, 2024.
\newblock Accessed: 2024-06-05.

\bibitem{guo2022longt5}
Mandy Guo, Joshua Ainslie, David Uthus, Santiago Ontanon, Jianmo Ni, Yun-Hsuan
  Sung, and Yinfei Yang.
\newblock {L}ong{T}5: {E}fficient text-to-text transformer for long sequences.
\newblock In {\em Findings of the Association for Computational Linguistics:
  NAACL 2022}, pages 724--736, Seattle, United States, 2022. Association for
  Computational Linguistics.

\bibitem{Guu2020REALMRL}
Kelvin Guu, Kenton Lee, Zora Tung, Panupong Pasupat, and Ming-Wei Chang.
\newblock Realm: retrieval-augmented language model pre-training.
\newblock In {\em Proceedings of the 37th International Conference on Machine
  Learning}, ICML'20. JMLR.org, 2020.

\bibitem{Hsieh2024RULERWT}
Cheng-Ping Hsieh, Simeng Sun, Samuel Kriman, Shantanu Acharya, Dima Rekesh, Fei
  Jia, and Boris Ginsburg.
\newblock Ruler: What's the real context size of your long-context language
  models?
\newblock {\em ArXiv}, 2024.

\bibitem{Hu2023OpendomainVE}
Hexiang Hu, Yi~Luan, Yang Chen, Urvashi Khandelwal, Mandar Joshi, Kenton Lee,
  Kristina Toutanova, and Ming-Wei Chang.
\newblock Open-domain visual entity recognition: Towards recognizing millions
  of wikipedia entities.
\newblock {\em 2023 IEEE/CVF International Conference on Computer Vision
  (ICCV)}, pages 12031--12041, 2023.

\bibitem{KamradtNeedle}
Greg Kamradt.
\newblock Needle in a haystack - pressure testing llms, 2023.

\bibitem{Karpukhin2020DensePR}
Vladimir Karpukhin, Barlas Oguz, Sewon Min, Patrick Lewis, Ledell Wu, Sergey
  Edunov, Danqi Chen, and Wen-tau Yih.
\newblock Dense passage retrieval for open-domain question answering.
\newblock In {\em Proceedings of the 2020 Conference on Empirical Methods in
  Natural Language Processing (EMNLP)}, pages 6769--6781, Online, 2020.
  Association for Computational Linguistics.

\bibitem{Kojima2022LargeLM}
Takeshi Kojima, Shixiang~(Shane) Gu, Machel Reid, Yutaka Matsuo, and Yusuke
  Iwasawa.
\newblock Large language models are zero-shot reasoners.
\newblock In S.~Koyejo, S.~Mohamed, A.~Agarwal, D.~Belgrave, K.~Cho, and A.~Oh,
  editors, {\em Advances in Neural Information Processing Systems}, volume~35,
  pages 22199--22213. Curran Associates, Inc., 2022.

\bibitem{kudo2018sentencepiece}
Taku Kudo and John Richardson.
\newblock {S}entence{P}iece: A simple and language independent subword
  tokenizer and detokenizer for neural text processing.
\newblock In {\em Proceedings of the 2018 Conference on Empirical Methods in
  Natural Language Processing: System Demonstrations}, pages 66--71, Brussels,
  Belgium, 2018. Association for Computational Linguistics.

\bibitem{Kwiatkowski2019NaturalQA}
Tom Kwiatkowski, Jennimaria Palomaki, Olivia Redfield, Michael Collins, Ankur
  Parikh, Chris Alberti, Danielle Epstein, Illia Polosukhin, Jacob Devlin,
  Kenton Lee, Kristina Toutanova, Llion Jones, Matthew Kelcey, Ming-Wei Chang,
  Andrew~M. Dai, Jakob Uszkoreit, Quoc Le, and Slav Petrov.
\newblock Natural questions: A benchmark for question answering research.
\newblock {\em Transactions of the Association for Computational Linguistics},
  7:452--466, 2019.

\bibitem{Lee2021YouON}
Haejun Lee, Akhil Kedia, Jongwon Lee, Ashwin Paranjape, Christopher Manning,
  and Kyoung-Gu Woo.
\newblock You only need one model for open-domain question answering.
\newblock In {\em Proceedings of the 2022 Conference on Empirical Methods in
  Natural Language Processing}, pages 3047--3060, Abu Dhabi, United Arab
  Emirates, 2022. Association for Computational Linguistics.

\bibitem{Lee2024GeckoVT}
Jinhyuk Lee, Zhuyun Dai, Xiaoqi Ren, Blair Chen, Daniel Cer, Jeremy~R. Cole,
  Kai Hui, Michael Boratko, Rajvi Kapadia, Wen Ding, Yi~Luan, Sai Meher~Karthik
  Duddu, Gustavo~Hern{\'a}ndez Abrego, Weiqiang Shi, Nithi Gupta, Aditya
  Kusupati, Prateek Jain, Siddhartha~R. Jonnalagadda, Ming-Wei Chang, and
  Iftekhar Naim.
\newblock Gecko: Versatile text embeddings distilled from large language
  models.
\newblock {\em ArXiv preprint}, abs/2403.20327, 2024.

\bibitem{Levy2024SameTM}
Mosh Levy, Alon Jacoby, and Yoav Goldberg.
\newblock Same task, more tokens: the impact of input length on the reasoning
  performance of large language models.
\newblock {\em ArXiv preprint}, abs/2402.14848, 2024.

\bibitem{Lewis2020QuestionAA}
Patrick Lewis, Pontus Stenetorp, and Sebastian Riedel.
\newblock Question and answer test-train overlap in open-domain question
  answering datasets.
\newblock In {\em Proceedings of the 16th Conference of the European Chapter of
  the Association for Computational Linguistics: Main Volume}, pages
  1000--1008, Online, 2021. Association for Computational Linguistics.

\bibitem{Lewis2020RetrievalAugmentedGF}
Patrick S.~H. Lewis, Ethan Perez, Aleksandra Piktus, Fabio Petroni, Vladimir
  Karpukhin, Naman Goyal, Heinrich K{\"{u}}ttler, Mike Lewis, Wen{-}tau Yih,
  Tim Rockt{\"{a}}schel, Sebastian Riedel, and Douwe Kiela.
\newblock Retrieval-augmented generation for knowledge-intensive {NLP} tasks.
\newblock In Hugo Larochelle, Marc'Aurelio Ranzato, Raia Hadsell,
  Maria{-}Florina Balcan, and Hsuan{-}Tien Lin, editors, {\em Advances in
  Neural Information Processing Systems 33: Annual Conference on Neural
  Information Processing Systems 2020, NeurIPS 2020, December 6-12, 2020,
  virtual}, 2020.

\bibitem{Li2024LongcontextLS}
Tianle Li, Ge~Zhang, Quy~Duc Do, Xiang Yue, and Wenhu Chen.
\newblock Long-context llms struggle with long in-context learning.
\newblock {\em ArXiv preprint}, abs/2404.02060, 2024.

\bibitem{mscoco}
Tsung-Yi Lin, Michael Maire, Serge Belongie, James Hays, Pietro Perona, Deva
  Ramanan, Piotr Doll{\'a}r, and C~Lawrence Zitnick.
\newblock Microsoft coco: Common objects in context.
\newblock In {\em Computer Vision--ECCV 2014: 13th European Conference, Zurich,
  Switzerland, September 6-12, 2014, Proceedings, Part V 13}, pages 740--755.
  Springer, 2014.

\bibitem{Liu2023LostIT}
Nelson~F. Liu, Kevin Lin, John Hewitt, Ashwin Paranjape, Michele Bevilacqua,
  Fabio Petroni, and Percy Liang.
\newblock Lost in the middle: How language models use long contexts.
\newblock {\em Transactions of the Association for Computational Linguistics},
  12:157--173, 2023.

\bibitem{Longpre2021EntityBasedKC}
Shayne Longpre, Kartik Perisetla, Anthony Chen, Nikhil Ramesh, Chris DuBois,
  and Sameer Singh.
\newblock Entity-based knowledge conflicts in question answering.
\newblock In {\em Proceedings of the 2021 Conference on Empirical Methods in
  Natural Language Processing}, pages 7052--7063, Online and Punta Cana,
  Dominican Republic, 2021. Association for Computational Linguistics.

\bibitem{Luo2023DrICLDI}
Man Luo, Xin Xu, Zhuyun Dai, Panupong Pasupat, Mehran Kazemi, Chitta Baral,
  Vaiva Imbrasaite, and Vincent Zhao.
\newblock Dr.icl: Demonstration-retrieved in-context learning.
\newblock {\em ArXiv preprint}, abs/2305.14128, 2023.

\bibitem{Maia2018WWW18OC}
Macedo Maia, Siegfried Handschuh, Andr{\'e} Freitas, Brian Davis, Ross
  McDermott, Manel Zarrouk, and Alexandra Balahur.
\newblock Www'18 open challenge: Financial opinion mining and question
  answering.
\newblock {\em Companion Proceedings of the The Web Conference 2018}, 2018.

\bibitem{Malaviya2023QUESTAR}
Chaitanya Malaviya, Peter Shaw, Ming-Wei Chang, Kenton Lee, and Kristina
  Toutanova.
\newblock {QUEST}: A retrieval dataset of entity-seeking queries with implicit
  set operations.
\newblock In Anna Rogers, Jordan Boyd-Graber, and Naoaki Okazaki, editors, {\em
  Proceedings of the 61st Annual Meeting of the Association for Computational
  Linguistics (Volume 1: Long Papers)}, pages 14032--14047, Toronto, Canada,
  July 2023. Association for Computational Linguistics.

\bibitem{min2021joint}
Sewon Min, Kenton Lee, Ming-Wei Chang, Kristina Toutanova, and Hannaneh
  Hajishirzi.
\newblock Joint passage ranking for diverse multi-answer retrieval.
\newblock In {\em Proceedings of the 2021 Conference on Empirical Methods in
  Natural Language Processing}, pages 6997--7008, Online and Punta Cana,
  Dominican Republic, 2021. Association for Computational Linguistics.

\bibitem{ni2022large}
Jianmo Ni, Chen Qu, Jing Lu, Zhuyun Dai, Gustavo Hernandez~Abrego, Ji~Ma,
  Vincent Zhao, Yi~Luan, Keith Hall, Ming-Wei Chang, and Yinfei Yang.
\newblock Large dual encoders are generalizable retrievers.
\newblock In {\em Proceedings of the 2022 Conference on Empirical Methods in
  Natural Language Processing}, pages 9844--9855, Abu Dhabi, United Arab
  Emirates, 2022. Association for Computational Linguistics.

\bibitem{Nye2021ShowYW}
Maxwell Nye, Anders~Johan Andreassen, Guy Gur-Ari, Henryk Michalewski, Jacob
  Austin, David Bieber, David Dohan, Aitor Lewkowycz, Maarten Bosma, David
  Luan, Charles Sutton, and Augustus Odena.
\newblock Show your work: Scratchpads for intermediate computation with
  language models.
\newblock In {\em Deep Learning for Code Workshop}, 2022.

\bibitem{Achiam2023GPT4TR}
OpenAI.
\newblock Gpt-4 technical report.
\newblock {\em ArXiv}, 2023.

\bibitem{Perez2020UnsupervisedQD}
Ethan Perez, Patrick Lewis, Wen-tau Yih, Kyunghyun Cho, and Douwe Kiela.
\newblock Unsupervised question decomposition for question answering.
\newblock In {\em Proceedings of the 2020 Conference on Empirical Methods in
  Natural Language Processing (EMNLP)}, pages 8864--8880, Online, 2020.
  Association for Computational Linguistics.

\bibitem{clip}
Alec Radford, Jong~Wook Kim, Chris Hallacy, Aditya Ramesh, Gabriel Goh,
  Sandhini Agarwal, Girish Sastry, Amanda Askell, Pamela Mishkin, Jack Clark,
  Gretchen Krueger, and Ilya Sutskever.
\newblock Learning transferable visual models from natural language
  supervision.
\newblock In Marina Meila and Tong Zhang, editors, {\em Proceedings of the 38th
  International Conference on Machine Learning, {ICML} 2021, 18-24 July 2021,
  Virtual Event}, volume 139 of {\em Proceedings of Machine Learning Research},
  pages 8748--8763. {PMLR}, 2021.

\bibitem{springer2024repetition}
Jacob~Mitchell Springer, Suhas Kotha, Daniel Fried, Graham Neubig, and Aditi
  Raghunathan.
\newblock Repetition improves language model embeddings.
\newblock {\em ArXiv preprint}, abs/2402.15449, 2024.

\bibitem{Suzgun2022ChallengingBT}
Mirac Suzgun, Nathan Scales, Nathanael Sch{\"a}rli, Sebastian Gehrmann, Yi~Tay,
  Hyung~Won Chung, Aakanksha Chowdhery, Quoc Le, Ed~Chi, Denny Zhou, and Jason
  Wei.
\newblock Challenging {BIG}-bench tasks and whether chain-of-thought can solve
  them.
\newblock In Anna Rogers, Jordan Boyd-Graber, and Naoaki Okazaki, editors, {\em
  Findings of the Association for Computational Linguistics: ACL 2023}, pages
  13003--13051, Toronto, Canada, July 2023. Association for Computational
  Linguistics.

\bibitem{Tay2020LongRA}
Yi~Tay, Mostafa Dehghani, Samira Abnar, Yikang Shen, Dara Bahri, Philip Pham,
  Jinfeng Rao, Liu Yang, Sebastian Ruder, and Donald Metzler.
\newblock Long range arena : {A} benchmark for efficient transformers.
\newblock In {\em 9th International Conference on Learning Representations,
  {ICLR} 2021, Virtual Event, Austria, May 3-7, 2021}. OpenReview.net, 2021.

\bibitem{tay2022transformer}
Yi~Tay, Vinh Tran, Mostafa Dehghani, Jianmo Ni, Dara Bahri, Harsh Mehta, Zhen
  Qin, Kai Hui, Zhe Zhao, Jai Gupta, et~al.
\newblock Transformer memory as a differentiable search index.
\newblock {\em Advances in Neural Information Processing Systems},
  35:21831--21843, 2022.

\bibitem{Reid2024Gemini1U}
Gemini Team.
\newblock Gemini 1.5: Unlocking multimodal understanding across millions of
  tokens of context.
\newblock {\em ArXiv preprint}, abs/2403.05530, 2024.

\bibitem{Thakur2021BEIRAH}
Nandan Thakur, Nils Reimers, Andreas R\"{u}ckl\'{e}, Abhishek Srivastava, and
  Iryna Gurevych.
\newblock Beir: A heterogeneous benchmark for zero-shot evaluation of
  information retrieval models.
\newblock In J.~Vanschoren and S.~Yeung, editors, {\em Proceedings of the
  Neural Information Processing Systems Track on Datasets and Benchmarks},
  volume~1, 2021.

\bibitem{Thorne2018FEVERAL}
James Thorne, Andreas Vlachos, Christos Christodoulopoulos, and Arpit Mittal.
\newblock {FEVER}: a large-scale dataset for fact extraction and
  {VER}ification.
\newblock In {\em Proceedings of the 2018 Conference of the North {A}merican
  Chapter of the Association for Computational Linguistics: Human Language
  Technologies, Volume 1 (Long Papers)}, pages 809--819, New Orleans,
  Louisiana, 2018. Association for Computational Linguistics.

\bibitem{Trivedi2021MM}
Harsh Trivedi, Niranjan Balasubramanian, Tushar Khot, and Ashish Sabharwal.
\newblock {M}u{S}i{Q}ue: Multihop questions via single-hop question
  composition.
\newblock {\em Transactions of the Association for Computational Linguistics},
  10:539--554, 2022.

\bibitem{Wachsmuth2018RetrievalOT}
Henning Wachsmuth, Shahbaz Syed, and Benno Stein.
\newblock Retrieval of the best counterargument without prior topic knowledge.
\newblock In {\em Proceedings of the 56th Annual Meeting of the Association for
  Computational Linguistics (Volume 1: Long Papers)}, pages 241--251,
  Melbourne, Australia, 2018. Association for Computational Linguistics.

\bibitem{Wadden2020FactOF}
David Wadden, Shanchuan Lin, Kyle Lo, Lucy~Lu Wang, Madeleine van Zuylen, Arman
  Cohan, and Hannaneh Hajishirzi.
\newblock Fact or fiction: Verifying scientific claims.
\newblock In {\em Proceedings of the 2020 Conference on Empirical Methods in
  Natural Language Processing (EMNLP)}, pages 7534--7550, Online, 2020.
  Association for Computational Linguistics.

\bibitem{wang2024adaleval}
Chonghua Wang, Haodong Duan, Songyang Zhang, Dahua Lin, and Kai Chen.
\newblock Ada-leval: Evaluating long-context llms with length-adaptable
  benchmarks, 2024.

\bibitem{Wei2021FinetunedLM}
Jason Wei, Maarten Bosma, Vincent~Y. Zhao, Kelvin Guu, Adams~Wei Yu, Brian
  Lester, Nan Du, Andrew~M. Dai, and Quoc~V. Le.
\newblock Finetuned language models are zero-shot learners.
\newblock In {\em The Tenth International Conference on Learning
  Representations, {ICLR} 2022, Virtual Event, April 25-29, 2022}.
  OpenReview.net, 2022.

\bibitem{Wei2022ChainOT}
Jason Wei, Xuezhi Wang, Dale Schuurmans, Maarten Bosma, Ed~Huai hsin Chi,
  F.~Xia, Quoc Le, and Denny Zhou.
\newblock Chain of thought prompting elicits reasoning in large language
  models.
\newblock {\em ArXiv preprint}, abs/2201.11903, 2022.

\bibitem{Xiong2020ApproximateNN}
Lee Xiong, Chenyan Xiong, Ye~Li, Kwok{-}Fung Tang, Jialin Liu, Paul~N. Bennett,
  Junaid Ahmed, and Arnold Overwijk.
\newblock Approximate nearest neighbor negative contrastive learning for dense
  text retrieval.
\newblock In {\em 9th International Conference on Learning Representations,
  {ICLR} 2021, Virtual Event, Austria, May 3-7, 2021}. OpenReview.net, 2021.

\bibitem{Xu2016MSRVTTAL}
Jun Xu, Tao Mei, Ting Yao, and Yong Rui.
\newblock {MSR-VTT:} {A} large video description dataset for bridging video and
  language.
\newblock In {\em 2016 {IEEE} Conference on Computer Vision and Pattern
  Recognition, {CVPR} 2016, Las Vegas, NV, USA, June 27-30, 2016}, pages
  5288--5296. {IEEE} Computer Society, 2016.

\bibitem{Yang2018HotpotQAAD}
Zhilin Yang, Peng Qi, Saizheng Zhang, Yoshua Bengio, William Cohen, Ruslan
  Salakhutdinov, and Christopher~D. Manning.
\newblock {H}otpot{QA}: A dataset for diverse, explainable multi-hop question
  answering.
\newblock In {\em Proceedings of the 2018 Conference on Empirical Methods in
  Natural Language Processing}, pages 2369--2380, Brussels, Belgium, 2018.
  Association for Computational Linguistics.

\bibitem{flickr}
Peter Young, Alice Lai, Micah Hodosh, and Julia Hockenmaier.
\newblock From image descriptions to visual denotations: New similarity metrics
  for semantic inference over event descriptions.
\newblock {\em Transactions of the Association for Computational Linguistics},
  2:67--78, 2014.

\bibitem{yu2020dialogue}
Dian Yu, Kai Sun, Claire Cardie, and Dong Yu.
\newblock Dialogue-based relation extraction.
\newblock In {\em Proceedings of the 58th Annual Meeting of the Association for
  Computational Linguistics}, pages 4927--4940, Online, 2020. Association for
  Computational Linguistics.

\bibitem{Yu2020DialogueBasedRE}
Dian Yu, Kai Sun, Claire Cardie, and Dong Yu.
\newblock Dialogue-based relation extraction.
\newblock In {\em Proceedings of the 58th Annual Meeting of the Association for
  Computational Linguistics}, pages 4927--4940, Online, 2020. Association for
  Computational Linguistics.

\bibitem{Yu2018SpiderAL}
Tao Yu, Rui Zhang, Kai Yang, Michihiro Yasunaga, Dongxu Wang, Zifan Li, James
  Ma, Irene Li, Qingning Yao, Shanelle Roman, Zilin Zhang, and Dragomir Radev.
\newblock {S}pider: A large-scale human-labeled dataset for complex and
  cross-domain semantic parsing and text-to-{SQL} task.
\newblock In {\em Proceedings of the 2018 Conference on Empirical Methods in
  Natural Language Processing}, pages 3911--3921, Brussels, Belgium, 2018.
  Association for Computational Linguistics.

\bibitem{Yu2019SParCCS}
Tao Yu, Rui Zhang, Michihiro Yasunaga, Yi~Chern Tan, Xi~Victoria Lin, Suyi Li,
  Heyang Er, Irene Li, Bo~Pang, Tao Chen, Emily Ji, Shreya Dixit, David
  Proctor, Sungrok Shim, Jonathan Kraft, Vincent Zhang, Caiming Xiong, Richard
  Socher, and Dragomir Radev.
\newblock {SP}ar{C}: Cross-domain semantic parsing in context.
\newblock In {\em Proceedings of the 57th Annual Meeting of the Association for
  Computational Linguistics}, pages 4511--4523, Florence, Italy, 2019.
  Association for Computational Linguistics.

\bibitem{zhong2017seq2sql}
Victor Zhong, Caiming Xiong, and Richard Socher.
\newblock Seq2{SQL}: Generating structured queries from natural language using
  reinforcement learning.
\newblock {\em ArXiv preprint}, abs/1709.00103, 2017.

\end{thebibliography}
\newpage

\appendix

\clearpage

\newpage

\section{\benchmark{} Dataset Creation}
\label{appendix:dataset-selection}

\subsection{Dataset Selection}
    \paragraph{Text Retrieval \& RAG}
    We test single-document retrieval on a representative subset of the BEIR benchmark~\citep{Thakur2021BEIRAH}, prioritizing datasets with high-quality ground truth labels~\citep{Wachsmuth2018RetrievalOT,Thorne2018FEVERAL,Maia2018WWW18OC,Bajaj2016MSMA,Kwiatkowski2019NaturalQA, Wadden2020FactOF}.
    We also include TopiOCQA~\citep{Adlakha2021TopiOCQAOC}, which is a multi-turn conversational retrieval dataset.
    We measure performance on single-document retrieval using Recall@1.
    Additionally, we test multi-document retrieval on HotPotQA~\citep{Yang2018HotpotQAAD}, MuSiQue~\citep{Trivedi2021MM}, QAMPARI~\citep{Malaviya2023QUESTAR}, where a set of documents must be retrieved to answer the query.
    The evaluation metric for multi-document retrieval is MRecall@$k$~\citep{min2021joint}, which gives a score of 1.0 if all $k$ gold set items are retrieved in top-$k$ and 0.0 otherwise.
    When creating the \benchmark{} version of the multi-document retrieval datasets, we limit the number of relevant documents per query to $k=2$, 5, 5, and 3 for HotPotQA, MuSiQue, QAMPARI, and QUEST, respectively, and the corresponding $k$'s are used for MRecall@$k$ (e.g. HotPotQA uses MRecall@2).
    
    Our RAG task contains subsets of retrieval datasets, which have phrase-level answer annotations: Natural Questions, TopiOCQA, HotPotQA, MuSiQue, QAMPARI, and QUEST.
    We use subspan exact match (EM)~\citep{10.1162/tacl_a_00667} for evaluating performance of all the datasets.
    In case of multi-answer datasets (i.e. QAMPARI, QUEST), we first match predicted answers to gold standard answers based on whether they overlap~\citep{dua-etal-2019-drop} via linear sum assignment algorithm. We then give full credit if every gold answer has a perfect match with aligned predicted answers.
    
    \paragraph{Visual Retrieval}
    We employ four diverse visual benchmarks: Flickr30k~\citep{flickr} and MSCOCO~\citep{mscoco} for text-to-image retrieval; MSR-VTT~\citep{Xu2016MSRVTTAL} for text-to-video retrieval (sampling 3 frames per video); and OVEN~\citep{Hu2023OpendomainVE} using the entity split for image-text retrieval where both queries and retrieval targets consist of image-text pairs.
    All images are resized to 512x512 and performance is assessed using Recall@1 for all datasets.
    
    \paragraph{Audio Retrieval}
    We utilize a subset of the multilingual FLEURS dataset~\citep{Conneau2022FLEURSFL}, focusing on the five most spoken languages\footnote{\url{https://en.wikipedia.org/wiki/List_of_languages_by_total_number_of_speakers}}: English (en), Hindi (hi), Chinese (zh), Spanish (es), and French (fr).
    Recall@1 is employed as the evaluation metric, given the single gold target.
    
    \paragraph{SQL}
    We evaluate SQL-like reasoning on Spider, a single-turn text-to-SQL dataset \cite{Yu2018SpiderAL}, and SparC, its multi-turn variant \cite{Yu2019SParCCS}.
    The input contains the database tables serialized as CSV and the natural language question.
    The model is allowed to perform reasoning in natural language before giving the final answer,
    which must be formatted in a Markdown code block.
    The extracted answers are evaluated against the execution results of the gold SQL queries.
    For SparC, the multi-turn questions are provided one-by-one in a conversational format, and credit is awarded only when the answers of all steps are correct.
    
    \paragraph{Many-shot ICL}
    We investigate \lclms{}' many-shot ICL capabilities by repurposing datasets from Big Bench Hard (BBH)~\citep{Srivastava2022BeyondTI,Suzgun2022ChallengingBT} and LongICLBench (LIB)~\citep{Yu2020DialogueBasedRE,Li2024LongcontextLS} to fit a many-shot ICL setting, focusing on multi-class classification tasks.
    The first set of datasets is drawn from Big-Bench Hard and includes: \texttt{date\_understanding} (BBH-date), \texttt{salient\_error\_translation\_detection} (BBH-salient), \texttt{tracking\_shuffled\_objects\_seven\_objects} (BBH-tracking7), and \texttt{web\_of\_lies} (BBH-web), each with up to 150 examples for prompting and up to 7 classes.
    Unlike other \benchmark{} tasks, the full corpus fits within 32k tokens which leads us to also create variants from 2k to 32k context lengths.
    We use accuracy as our metric for Big Bench Hard.
    We also evaluate with \dialogue{}~\citep{Yu2020DialogueBasedRE}, a dialogue-based relation classification dataset with 36 relation labels.
    We follow the \LiclBench{} format but use accuracy as our metric.

\newpage
\section{Datasets Processing Details}\label{appendix:dataset_processing}

\paragraph{Content Filtering}
The language model APIs often block inputs with potentially harmful contents. When creating \benchmark{}, we tried to remove such contents from textual and visual inputs. Our filtering was done using a classifier as well as a keyword-based filtering. Despite our best effort, some API calls still refused to provide answers, which we treated as incorrect in our evaluation.

\paragraph{Tokenization}
To measure the size of a corpus, we count the number of tokens returned by the SentencePiece tokenizer~\citep{kudo2018sentencepiece}. 

\paragraph{Links to Dataset Sources}
\benchmark{} repurposes existing datasets for evaluating \lclms{}.
Here are the links to the original datasets used in \benchmark{}.
\begin{itemize}
    \item Text Retrieval - BEIR~\citep{Thakur2021BEIRAH} (ArguAna~\citep{Wachsmuth2018RetrievalOT}, FEVER~\citep{Thorne2018FEVERAL}, FIQA~\citep{Maia2018WWW18OC}, MS MARCO~\citep{Bajaj2016MSMA}, NQ~\citep{Kwiatkowski2019NaturalQA}, Quora, SciFact~\citep{Wadden2020FactOF}, Touché-2020~\citep{Bondarenko2020OverviewOT}, HotPotQA~\citep{Yang2018HotpotQAAD}): \url{https://github.com/beir-cellar/beir}
    \item Text Retrieval - TopiOCQA~\citep{Adlakha2021TopiOCQAOC}: \url{https://github.com/McGill-NLP/topiocqa}
    \item Text Retrieval - MuSiQue~\citep{Trivedi2021MM}: \url{https://allenai.org/data/musique}
    \item Text Retrieval - QAMPARI~\citep{Amouyal2022QAMPARIAB}: \url{https://github.com/samsam3232/qampari}
    \item Text Retrieval - QUEST~\citep{Malaviya2023QUESTAR}: \url{https://github.com/google-research/language/tree/master/language/quest}
    \item Visual Retrieval - Flickr30k~\citep{flickr}: \url{https://www.kaggle.com/datasets/hsankesara/flickr-image-dataset}
    \item Visual Retrieval - MS COCO~\citep{mscoco}: \url{https://cocodataset.org}
    \item Visual Retrieval - OVEN~\citep{Hu2023OpendomainVE}: \url{https://github.com/open-vision-language/oven}
    \item Visual Retrieval - MSR-VTT~\citep{Xu2016MSRVTTAL}: \url{https://cove.thecvf.com/datasets/839}
    \item Audio Retrieval - FLEURS~\citep{Conneau2022FLEURSFL}: \url{https://huggingface.co/datasets/google/fleurs}
    \item RAG - Same as Text Retrieval
    \item SQL - Spider~\citep{Yu2018SpiderAL}: \url{https://yale-lily.github.io/spider}
    \item SQL - SparC~\citep{Yu2019SParCCS}: \url{https://yale-lily.github.io/sparc}
    \item Many-Shot ICL - Big-Bench Hard~\citep{Srivastava2022BeyondTI, Suzgun2022ChallengingBT}: \url{https://github.com/suzgunmirac/BIG-Bench-Hard}
    \item Many-Shot ICL - LongICLBench~\citep{Yu2020DialogueBasedRE,Li2024LongcontextLS}: \url{ https://github.com/TIGER-AI-Lab/LongICLBench}
\end{itemize}

\newpage
\section{Detailed Statistics}\label{appendix:dataset_stats_detail}
In \Cref{tab:dataset_statistics_appendix}, we show detailed statistics of the \benchmark{} benchmark.
\begin{table}[h]
    \centering
    \setlength\tabcolsep{3pt}
     \setlength\extrarowheight{-1pt}
     \resizebox{0.95\linewidth}{!}{
    \begin{tabular}{clccc}
    \toprule
    \textbf{Task} & \textbf{Dataset} & \makecell{\textbf{\# Queries}\\(Few-shot / Development / Test)} & \makecell{\textbf{Supported}\\\textbf{Context Length}} & \textbf{\# Candidates}\\
    \midrule
    \textbf{\multirow{13}*{\makecell{Text\\Retrieval}}} & ArguAna & 5 / 10 / 100 & 32k / 128k / 1M & 123 / 531 / 3,891 \\
    & FEVER & 5 / 10 / 100 & 32k / 128k / 1M & 154 / 588 / 6,031 \\
    & FIQA &  5 / 10 / 100 & 32k / 128k / 1M & 148 / 531 / 4,471 \\
    & MS MARCO & 5 / 10 / 100 & 32k / 128k / 1M & 302 / 1,174 / 9,208 \\
    & NQ & 5 / 10 / 100 & 32k / 128k / 1M & 214 / 883 / 6,999 \\
    & Quora & 5 / 10 / 100 & 32k / 128k / 1M & 820 / 3,306 / 25,755\\
    & SciFact & 5 / 10 / 100 & 32k / 128k / 1M & 86 / 357 / 2,753 \\
    & Touché-2020 & 5 / 10 / 34 & 32k / 128k / 1M & 77 / 329 / 2,843 \\
    & TopiOCQA & 5 / 10 / 100 & 32k / 128k / 1M & 170 / 680 / 5,379 \\
    & HotPotQA & 5 / 10 / 100 & 32k / 128k / 1M & 319 / 1,222 / 10,005 \\
    & MuSiQue & 5 / 10 / 100 & 32k / 128k / 1M & 210 / 824 / 6,650 \\
    & QAMPARI & 5 / 10 / 100 & 32k / 128k / 1M & 186 / 755 / 5,878\\
    & QUEST & 5 / 10 / 100 & 32k / 128k / 1M & 87 / 328 / 2,858 \\
    \midrule

    \textbf{\multirow{4}*{\makecell{Visual \\Retrieval}}} & Flickr30k & 5 / 10 / 100 & 32k / 128k & 115 / 440 \\
    & MS COCO & 5 / 10 / 100 & 32k / 128k / 1M & 115 / 440 / 3,448 \\
    & OVEN  & 5 / 10 / 100 & 32k / 128k / 1M & 110 / 448 / 3475 \\
    & MSR-VTT & 5 / 10 / 100 & 32k / 128k / 1M & 35 / 140 / 1,101\\
    \midrule
    \textbf{\multirow{5}*{\makecell{Audio\\Retrieval}}} & FLEURS-en  & 5 / 10 / 100 & 32k / 128k & 104 / 428 \\
    & FLEURS-es  & 5 / 10 / 100 & 32k / 128k & 77 / 343 \\
    & FLEURS-fr  & 5 / 10 / 100 & 32k / 128k & 94 / 412 \\
    & FLEURS-hi  & 5 / 10 / 100 & 32k / 128k & 83 / 369 \\
    & FLEURS-zh  & 5 / 10 / 100 & 32k / 128k & 85 / 370\\
    \midrule
    \textbf{\multirow{6}*{\makecell{RAG}}}  & NQ & 5 / 10 / 100 & 32k / 128k / 1M & 214 / 883 / 6,999 \\
    & TopiOCQA & 5 / 10 / 100 & 32k / 128k / 1M & 170 / 680 / 5,379 \\
    & HotPotQA & 5 / 10 / 100 & 32k / 128k / 1M & 319 / 1,222 / 10,005 \\
    & MuSiQue & 5 / 10 / 100 & 32k / 128k / 1M & 210 / 824 / 6,650 \\
    & QAMPARI & 5 / 10 / 100 & 32k / 128k / 1M & 186 / 755 / 5,878\\
    & QUEST & 5 / 10 / 100 & 32k / 128k / 1M & 87 / 328 / 2,858 \\
    \midrule
    \textbf{\multirow{2}*{SQL}} & Spider & 1 / 10 / 100 & 32k / 128k / 1M & 1 / 1 / 1\\
    & SParC & 1 / 10 / 100 & 32k / 128k / 1M & 1 / 1 / 1 \\
    \midrule
    \textbf{\multirow{5}*{\makecell{Many-Shot\\ICL}}} & BBH-date  & - / 10 / 90 & 32k & 150 \\
    & BBH-salient  &  - / 10 / 90 & 32k & 104 \\
    & BBH-tracking7 & - / 10 / 90 & 32k & 123 \\
    & BBH-web & - / 10 / 90 & 32k & 150\\
    & LIB-dialogue &  - / 10 / 100 & 32k / 128k / 1M & 61 / 274 / 1,059 \\
    \bottomrule
    \\
    \end{tabular}
    }
    \caption{
        \label{tab:dataset_statistics_appendix}
        Tasks and datasets in the \benchmark{} benchmark.
        We show the number of queries per each split and and supported context lengths for each dataset.
    }
\end{table}

\newpage
\section{Prompt Design}\label{appendix:dataset_instructions}
\subsection{Dataset Instructions}
\begin{table}[h]
    \centering
    \setlength\tabcolsep{3pt}
    \setlength\extrarowheight{-1pt}
    \resizebox{1.0\linewidth}{!}{
    \begin{tabular}{cp{123mm}}
    \toprule
    \textbf{Dataset} & \textbf{Instruction} \\
    \midrule
    \multicolumn{2}{c}{\textbf{Text Retrieval}} \\
    \midrule
    \multirow{3}{*}{\makecell{ArguAna}} & You will be given a list of statements. You need to read carefully and understand all of them. Then you will be given a claim, and your goal is to find all statements from the list that can counterargue the claim. \\
    \midrule
    \multirow{3}{*}{\makecell{FEVER\\Scifact}} & You will be given a list of passages. You need to read carefully and understand all of them. Then you will be given a claim, and your goal is to find all passages from the list that can help verify the claim as true of false.\\
    \midrule
    \multirow{3}{*}{\makecell{FIQA\\MS MARCO\\NQ, TopiOCQA}} & You will be given a list of documents. You need to read carefully and understand all of them. Then you will be given a query, and your goal is to find all documents from the list that can help answer the query. \\
    \midrule
    \multirow{3}{*}{\makecell{Quora}} & You will be given a list of questions. You need to read carefully and understand all of them. Then you will be given a new question, and your goal is to find all questions from the list that are near duplicates of the new question. \\
    \midrule
    \multirow{3}{*}{\makecell{Touché-2020}} & You will be given a list of arguments. You need to read carefully and understand all of them. Then you will be given a controversial debating topic, and your goal is to find arguments from the list that's relevant to the topic. \\
    \midrule
    \multirow{3}{*}{\makecell{HotPotQA\\MuSiQue\\QAMPARI\\QUEST}} & You will be given a list of documents. You need to read carefully and understand all of them. Then you will be given a query that may require you to use 1 or more documents to find the answer. Your goal is to find all documents from the list that can help answer the query. \\
    \midrule
    \multicolumn{2}{c}{\textbf{Visual Retrieval}} \\
    \midrule
    \multirow{3}{*}{\makecell{Flickr30k\\MS COCO}} & You will be given a list of images. You need to carefully watch all of them. Then you will be given a new sentence, and your goal is to find most relevant image from the list for the given sentence. \\
    \midrule
    \multirow{4}{*}{\makecell{OVEN}}  & You will be given a list of Wikipedia entries which contains Wikipedia ID, Title and Description image. You need to carefully watch all of them. Then you will be given a input image and a question related to the image, and your goal is to find most relevant Wikipedia entry from the list that can be used to best answer the question. \\
    \midrule
    \multirow{4}{*}{\makecell{MSR-VTT}} & You will be given a list of videos which contains the video ID and video content (present as sequence of images, with timestamp in text). You need to carefully watch all of them. Then you will be given a text query, and your goal is to find most relevant video from the list that can best answer the question. \\
    \midrule
    \multicolumn{2}{c}{\textbf{Audio Retrieval}} \\
    \midrule
    \multirow{4}{*}{\makecell{FLEURS-*}}  & You will be given a list of audio which contains Audio ID and audio. You need to carefully listen all of them. Then you will be given a transcript, and your goal is to find most relevant audio from the list that matches the given transcript. Print out the Audio ID of the audio presented in the list. \\
    \midrule
    \multicolumn{2}{c}{\textbf{SQL}} \\
    \midrule
    \multirow{4}{*}{\makecell{Spider\\SparC}} & You will be given a list of tables. You need to read all of the rows of each table. Then you will be given a query, and your goal is to get the answer from the tables. Then format the answer into a list of lists. When formatting the answer into a list of lists, make sure you use the exact fields that are provided in the tables.\\
    \bottomrule
    \\
    \end{tabular}
    }
    \caption{
        \label{tab:dataset_instructions}
        Instructions used for each \benchmark{} dataset.
        We omit instructions for the RAG datasets, which are almost identical to text retrieval instructions.
        The ICL task does not use additional instructions, but only many-shot examples in their context.
    }
\end{table}

\newpage
\subsection{Chain-of-Thought Reasoning}\label{appendix:cot_instruction}

We used two types of chain-of-thought reasoning in the few-shot examples in the prompts: 
\begin{itemize}
    \item \textbf{Handwritten reasoning.} For multi-hop retrieval and RAG datasets (HotPotQA and MuSiQue), as well as SQL datasets (Spider and SparC), we manually wrote reasoning chains for the few-shot queries.
    \item \textbf{Relevant content.} For the rest of the retrieval and RAG datasets, we simply let the model generate the title (or the passage if the title is not available) of each gold document before generating the final answers. This helps explain why the final answers (IDs or short phrases) are the correct ones.
\end{itemize}
Many-shot ICL task did not use any chain-of-thought reasoning.
We also noticed that GPT-4o performed much better without chain-of-thought reasoning, so the GPT-4o results presented in this paper did not utilize the chain-of-thought.

\section{Positional Analysis Detailed Results}
\label{appendix:positional-analysis}

\begin{figure*}[h]
\begin{center}
\includegraphics[width=0.6\columnwidth]{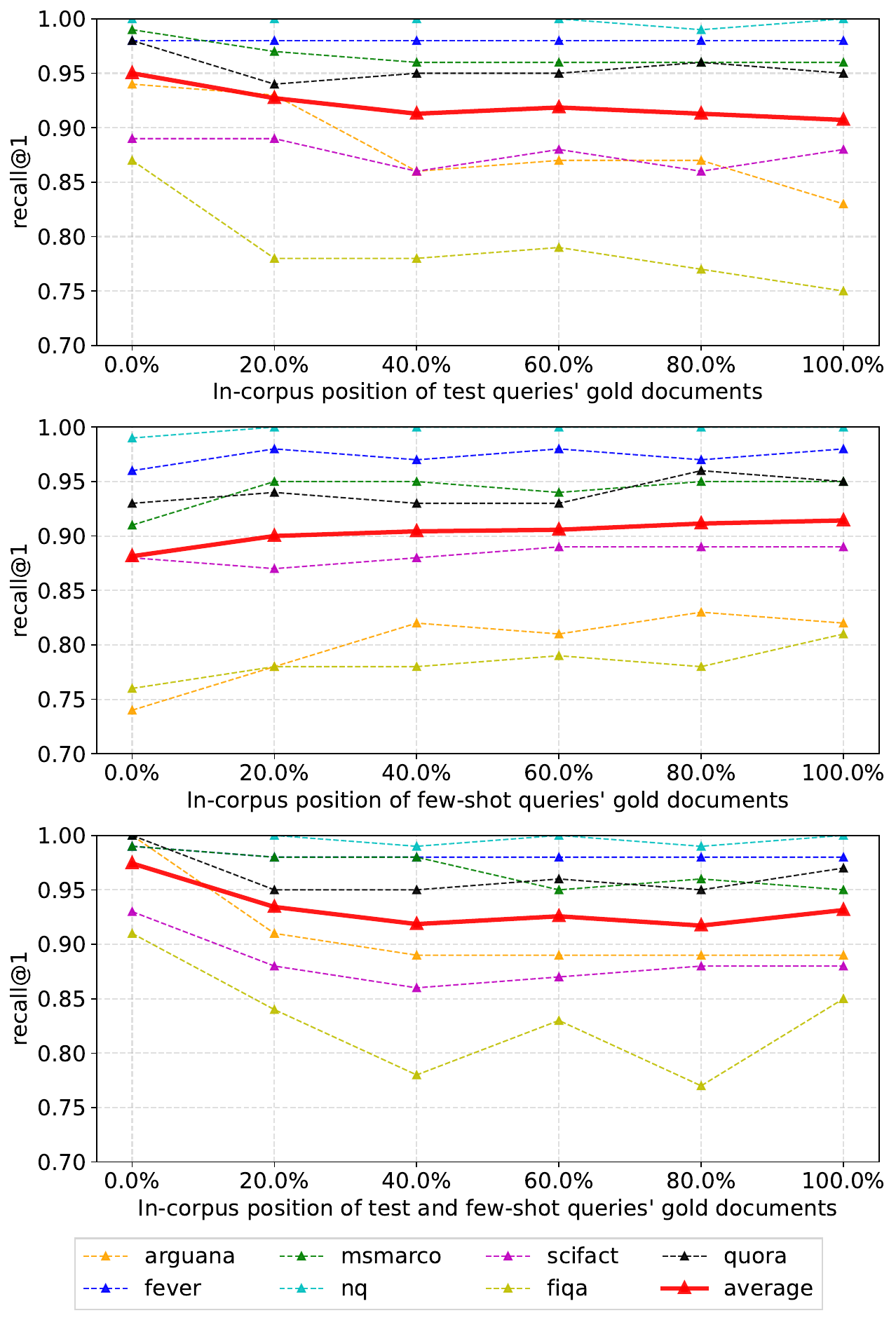}
\caption{Detailed metrics of the positional analysis, where we vary the position of gold documents of the test and few-shot queries. \textit{\textbf{Top:}} we vary the gold documents position of test queries within the corpus. \textit{\textbf{Middle:}} we vary the the gold documents position of few-shot queries within the corpus. \textit{\textbf{Bottom:}} we group the gold documents of test and few-shot queries together, and vary their position within the corpus. The average is shown in red.}
\label{fig:appendix_positional_analysis_details}
\end{center}
\end{figure*}

\clearpage
\section{Ablated Prompt Examples}
\label{appendix:ablated-prompt-examples}
\begin{figure}[h]
\centering
\includegraphics[width=1.0\columnwidth]{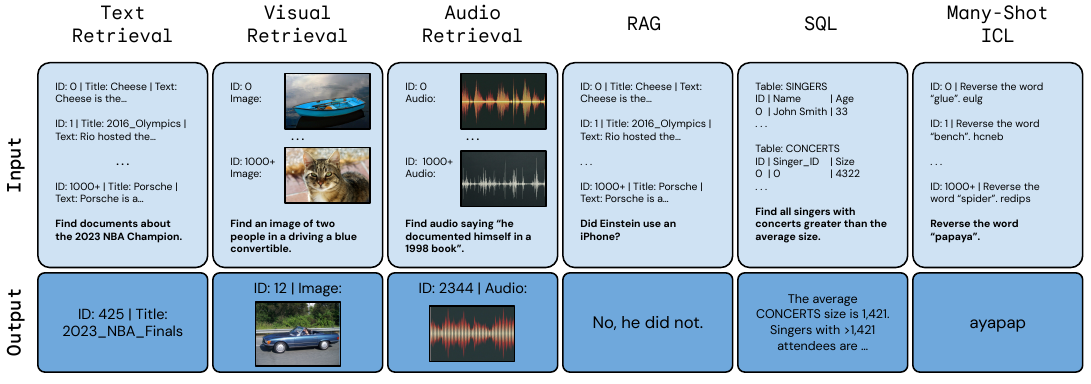}
\caption{
    \label{fig:dataset_examples}
    Examples of the task prompts in \benchmark{}.
    Each LCLM is expected to do in-context retrieval, reasoning, and many-shot learning on corpora up to millions of tokens.
}
\end{figure}

\begin{figure*}[h]
\begin{center}
\includegraphics[width=1\columnwidth]{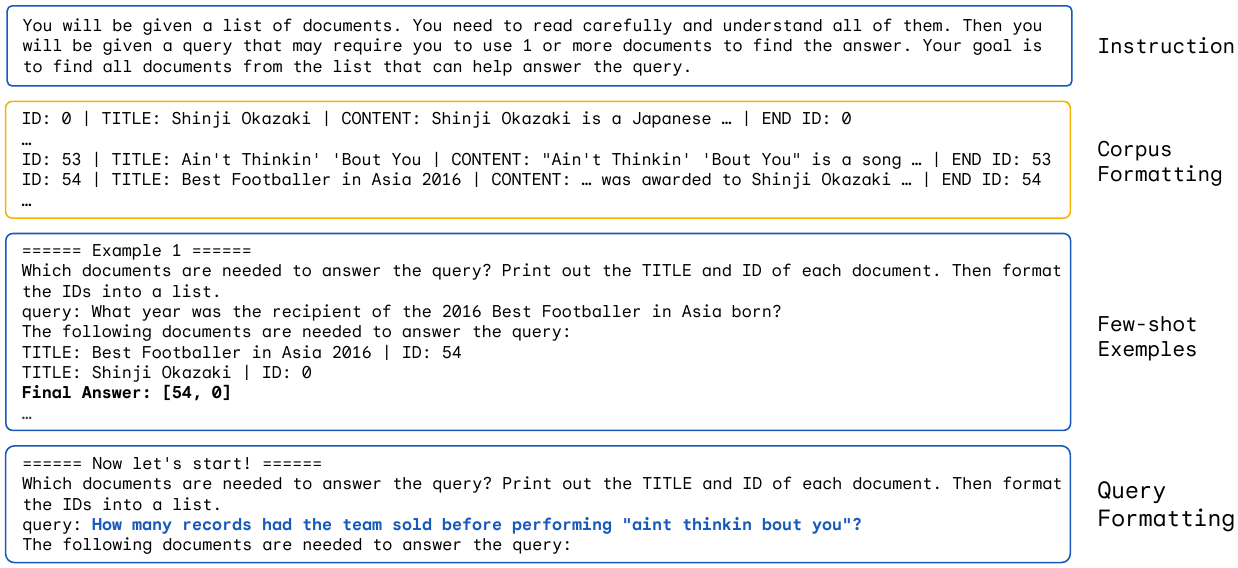}
\caption{Original CiC prompt for HotPotQA, a retrieval dataset in \benchmark{}. The prompt contains an instruction, a corpus, few-shot examples and a query.
}
\label{fig:appendix_ablation_cic}
\end{center}
\end{figure*}

\clearpage

\begin{figure*}[h]
\begin{center}
\includegraphics[width=1\columnwidth]{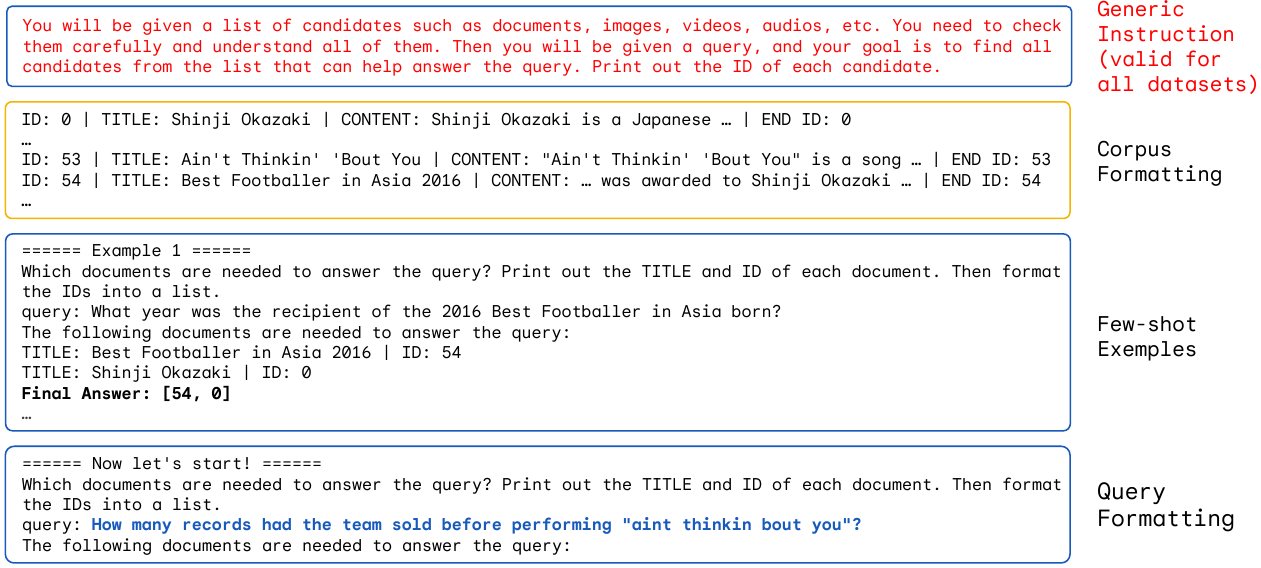}
\caption{Generic Instruction Ablation, with changes to the original CiC prompt in red. The instruction is changed to a generic one that applies to all tasks in \benchmark{}.
}
\label{fig:appendix_ablation_generic_instruction}
\end{center}
\end{figure*}

\begin{figure*}[h]
\begin{center}
\includegraphics[width=1\columnwidth]{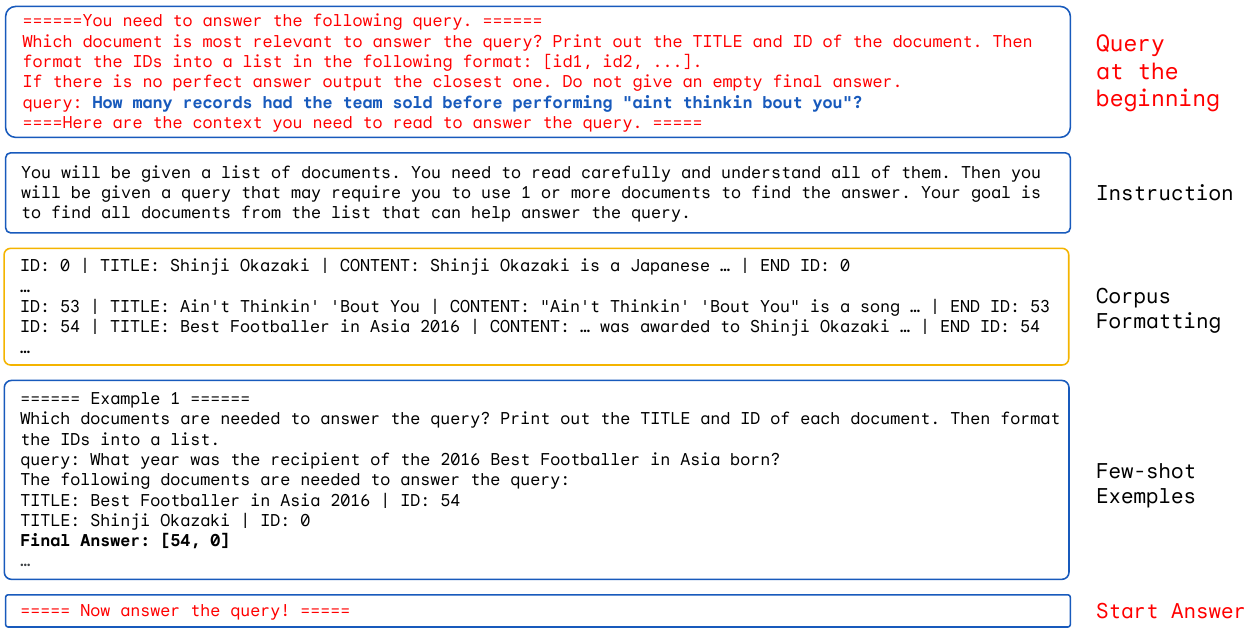}
\caption{Query at the Beginning Ablation, with changes to the original CiC prompt in red. The query is placed at the beginning instead of the end.}
\label{fig:appendix_ablation_query_beginning}
\end{center}
\end{figure*}

\clearpage

\begin{figure*}[h]
\begin{center}
\includegraphics[width=1\columnwidth]{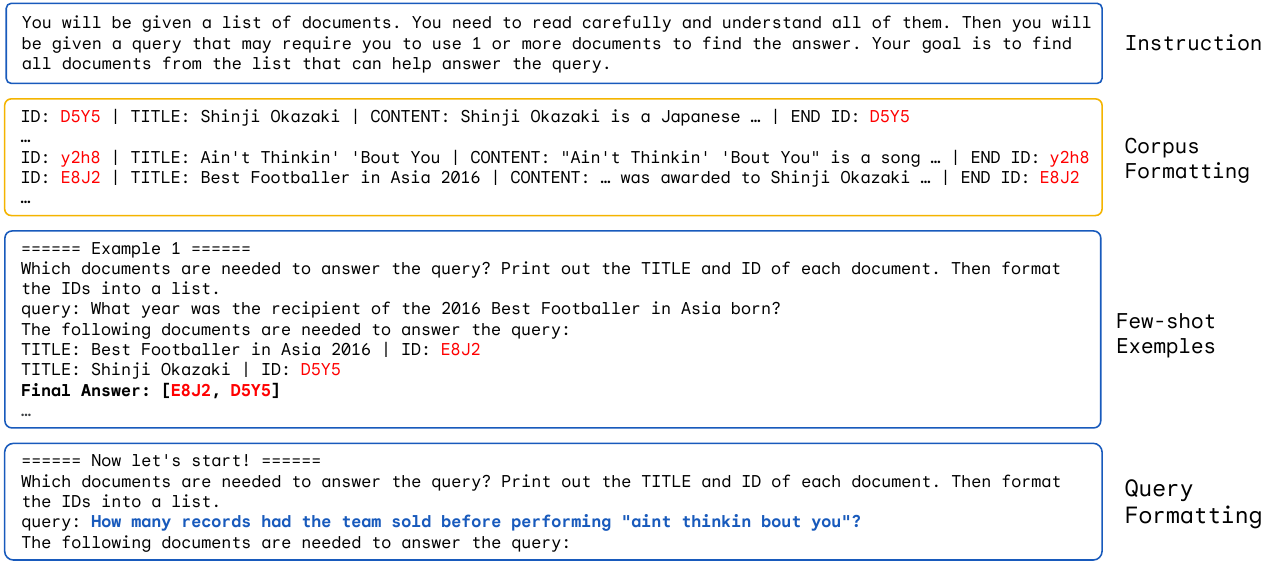}
\caption{Alphanumeric Document ID Ablation, with changes to the original CiC prompt in red. Instead of using sequential numeric document IDs, a unique random alphanumeric ID is generated with alternating ASCII letters and digits.}
\label{fig:appendix_ablation_alnum}
\end{center}
\end{figure*}

\begin{figure*}[h]
\begin{center}
\includegraphics[width=1\columnwidth]{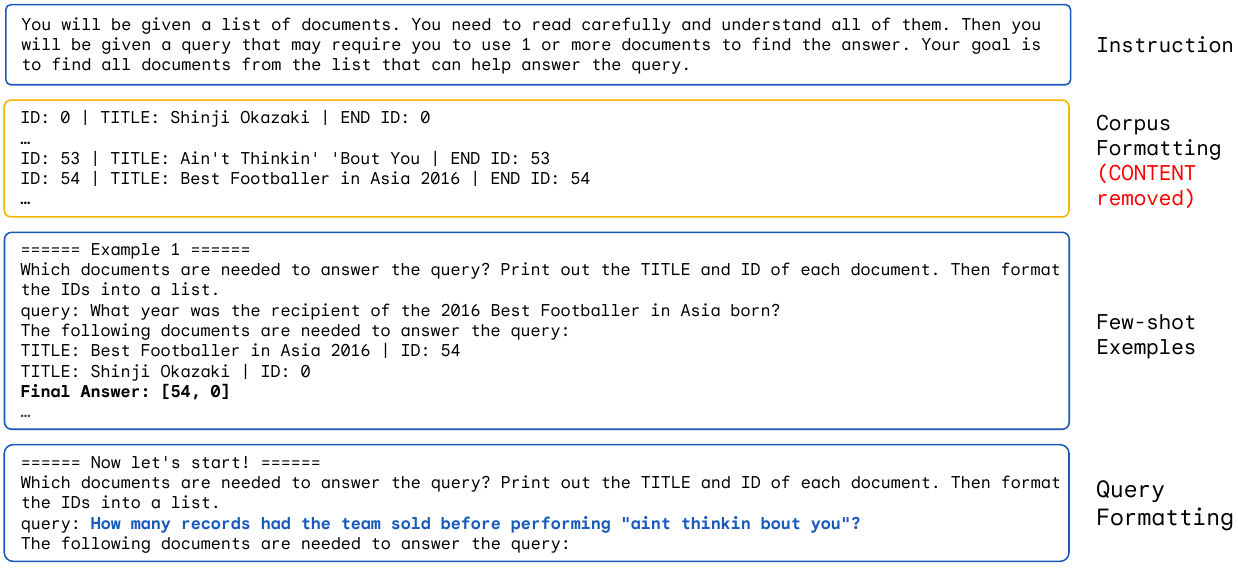}
\caption{Title Only Ablation, with changes to the original CiC prompt in red. In this ablation, the document content is removed, keeping only the document title.}
\label{fig:appendix_ablation_title_only}
\end{center}
\end{figure*}

\clearpage

\begin{figure*}[h]
\begin{center}
\includegraphics[width=1\columnwidth]{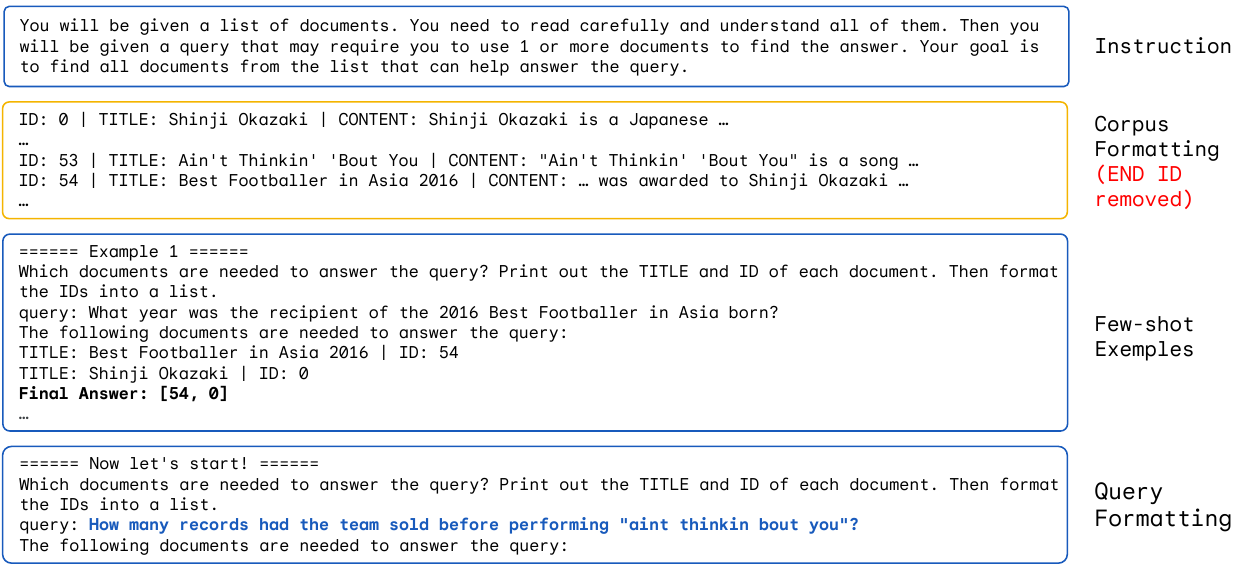}
\caption{ID Echo Ablation, with changes to the original CiC prompt in red. In this ablation, the ID is only mentioned at the beginning of each document, and we remove the ID echo at the end (e.g. "END ID:").}
\label{fig:appendix_ablation_corpus_echo}
\end{center}
\end{figure*}

\begin{figure*}[h]
\begin{center}
\includegraphics[width=1\columnwidth]{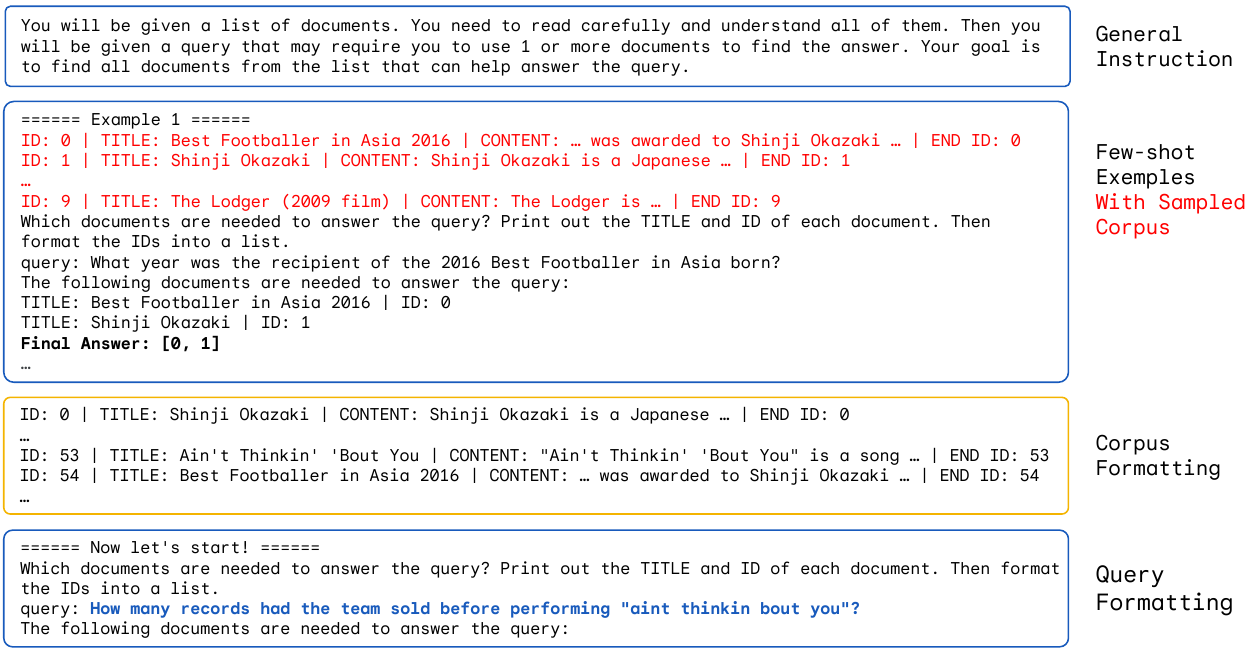}
\caption{Corpus in Each Few-shot Ablation, with changes to the original CiC prompt in red. In particular, in this ablation, each few-shot example contains a sampled corpus (10 documents), the full corpus is then given before the Query part of the prompt.}
\label{fig:appendix_ablation_corpus_in_few_shot}
\end{center}
\end{figure*}

\clearpage

\begin{figure*}[h]
\begin{center}
\includegraphics[width=1\columnwidth]{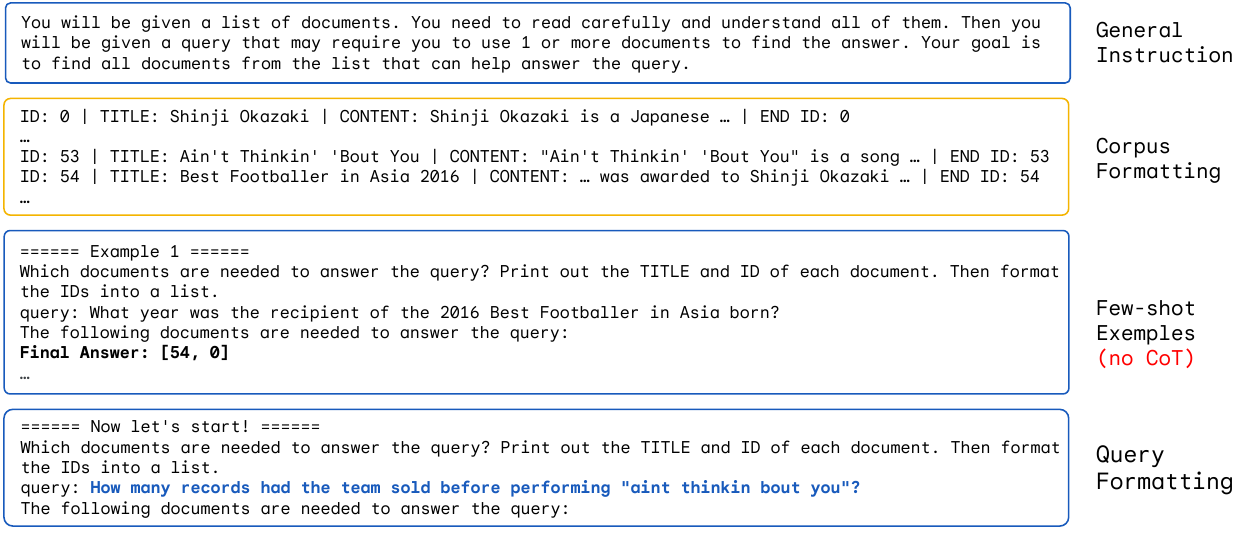}
\caption{No Chain-of-Thought Ablation, with changes to the original CiC prompt in red. In particular, in this ablation, we remove the chain-of-thought from the few-shot examples, instead opting to output the answer directly.}
\label{fig:appendix_ablation_no_cot}
\end{center}
\end{figure*}

\newpage
\section{Qualitative Analysis of Long-Context Wins and Losses}\label{appendix:qualitative}
\begin{table}[ht]
\centering
\footnotesize
\resizebox{1.0\linewidth}{!}{
    \begin{tabular}{p{15mm}p{120mm}}
        \toprule
        \multicolumn{2}{c}{Task = {\bf Retrieval}} \\
        \midrule
        \multicolumn{2}{c}{Dataset = \textbf{Natural Questions} 1M} \\
        \midrule
        \textbf{Input} & \texttt{\method{}-prompt}(query="when does monday night raw come on hulu")  \\
        \textbf{Output} & The following documents can help answer the query: \\
        & TITLE: WWE Raw | ID: 1023 \\
        & Final Answer: [\textquotesingle 1023\textquotesingle] \correct{} \\
        \midrule
        \textbf{Input} & \texttt{\method{}-prompt}(query="where does the synthesis of new dna from existing dna occurs")  \\
        \textbf{Output} & The following documents can help answer the query:\\
        & TITLE: DNA replication | ID: 3355 \\
        & Final Answer: [\textquotesingle 3355\textquotesingle] \correct{}\\
        \midrule
        \textbf{Input} & \texttt{\method{}-prompt}(query="jawaharlal nehru centre for advanced scientific research jakkur campus")  \\
        \textbf{Output} & TITLE: Jawaharlal Nehru Centre for Advanced Scientific Research | ID: 6911 \\
        & Final Answer: [\textquotesingle 6911\textquotesingle] \wrong{} // Incorrect ID (gold = \textquotesingle5691\textquotesingle) \\
        \midrule
        \multicolumn{2}{c}{Dataset = \textbf{HotPotQA} 128k (multi-hop)} \\
        \midrule
        \textbf{Input} & \texttt{\method{}-prompt}(query="Which internationl football team has had a player endorse the Nike Hypervenom football boot?")  \\
        \textbf{Output} & The following documents can help answer the query:\\
        & reasoning: The Nike Hypervenom is endorsed by players such as Marcus Rashford based on the document **ID:** 413 (title: Nike Hypervenom). \\
        & reasoning: Marcus Rashford plays for the England national team based on the document **ID:** 558 (title: Marcus Rashford). \\
        & Final Answer: [\textquotesingle 413\textquotesingle, \textquotesingle 558\textquotesingle] \correct{} \\
        \midrule
        \textbf{Input} & \texttt{\method{}-prompt}(query="What NHL team drafted Kaspars Astasenko in 1999 and won the Stanley Cup in 2003-04?")  \\
        \textbf{Output} & The following documents can help answer the query:\\
        & ID: 217 | TITLE: Kaspars Astašenko | CONTENT: Kaspars Astašenko (17 February 1975 – 20 November 2012) was a Latvian professional ice hockey player. Astašenko was born in Riga, Latvia. Astašenko was drafted by the Tampa Bay Lightning in the 1999 NHL Entry Draft, 127th overall. Astašenko played parts of two seasons in the National Hockey League with the Lightning. | END ID: 217\\
        & ID: 508 | TITLE: Tampa Bay Lightning | CONTENT: The Tampa Bay Lightning are a professional ice hockey team based in Tampa, Florida. It is a member of the Atlantic Division of the Eastern Conference of the National Hockey League (NHL). The Lightning have one Stanley Cup championship in their history, in 2003–04. The team is often referred to as the Bolts, and the nickname is used on the current third jersey. The Lightning plays home games in the Amalie Arena in Tampa. | END ID: 508\\
        & Final Answer: [\textquotesingle 217\textquotesingle, \textquotesingle 508\textquotesingle] \correct{} \\
        \midrule
        \textbf{Input} & \texttt{\method{}-prompt}(query="What drum machine was introduced in 1989 and allowed 32 notes to be played simultaneously?")  \\
        \textbf{Output} & The following documents can help answer the query:\\
        & ID: 773 | TITLE: Roland R-8 | CONTENT: The R-8 Human Rhythm Composer is an electronic drum machine introduced in 1989 by Roland Corporation, using PCM voices. ... The device has eight individual outputs, 32-voice polyphony, and four-part multitimbral MIDI. | END ID: 773\\
        & Final Answer: [\textquotesingle 773\textquotesingle] \wrong{} // Partially correct (\textquotesingle666\textquotesingle{} is missing) \\
        \bottomrule
        \\
    \end{tabular}
}
\caption{Qualitative examples showcasing the successes and failures of the \method{} prompting in the text retrieval task.
We use Gemini 1.5 Pro for the analysis.
For HotPotQA, we observed that some of the questions are actually answerable from a single document as illustrated in the last example.}
\label{tab:qual_examples_retrieval}
\end{table}

\begin{table}[t]
\centering
\footnotesize
\resizebox{1.0\linewidth}{!}{
    \begin{tabular}{p{15mm}p{120mm}}
        \toprule
        \multicolumn{2}{c}{Task = {\bf RAG}}  \\
        \midrule
        \multicolumn{2}{c}{{Dataset = \textbf{MuSiQue} 128k (multi-hop)}} \\
        \midrule
        \textbf{Input} & \texttt{\method{}-prompt}(query="Where do greyhound buses leave from in the city the performer of \underline{Born A Lion} formed in?")  \\
        \textbf{Output} & reasoning: \underline{Born A Lion} is the debut album by Canadian rock band \underline{Danko Jones}.\\
        & reasoning: \underline{Danko Jones} is a Canadian rock trio from \underline{Toronto, Ontario}.\\
        & reasoning: The \underline{Toronto Coach Terminal} is the central bus station for inter-city services in \underline{Toronto, Ontario}, Canada.\\
        & ID: 427 | TITLE: Born a Lion\\
        & ID: 383 | TITLE: Danko Jones\\
        & ID: 724 | TITLE: Toronto Coach Terminal\\
        & Final Answer: [\textquotesingle Toronto Coach Terminal\textquotesingle] \correct{} \\
        \midrule
        \textbf{Input} & \texttt{\method{}-prompt}(query="When did the rx 350 model of the luxury division of the company that built \underline{Daihatsu boon} change body style?") \\
        \textbf{Output} & reasoning: The \underline{Daihatsu Boon} is also sold as the \underline{Toyota Passo}.\\
        & reasoning: The \underline{Passo} is sold at "\underline{Toyota} Corolla Store" Japanese dealerships.\\
        & reasoning: Scion is a discontinued marque of \underline{Toyota}.\\
        & reasoning: The Lexus RX is a mid-size luxury crossover SUV sold since 1998 by Lexus, a luxury division of \underline{Toyota}.\\
        & reasoning: The first generation Lexus RX was introduced in 1998 and was produced until 2003.\\
        & reasoning: The second generation Lexus RX was introduced in 2003 and was produced until 2009.\\
        & reasoning: The third generation Lexus RX was introduced in 2008 and was produced until 2015.\\
        & reasoning: The fourth generation Lexus RX was introduced in 2015. \\
        & reasoning: The RX 350 model was introduced in the second generation Lexus RX in 2003.\\
        & Final Answer: [\textquotesingle 2003\textquotesingle, \textquotesingle 2009\textquotesingle, \textquotesingle 2015\textquotesingle ] \wrong{} // Incorrect (answer = \textquotesingle Sales began worldwide in April 2012\textquotesingle)\\
        \midrule
        \multicolumn{2}{c}{Dataset = {\textbf{QUEST} 128k (multi-answer)}} \\
        \midrule
        \textbf{Input} & \texttt{\method{}-prompt}(query="History books about \underline{Malaysia} or the \underline{Qing dynasty} or about \underline{South Korea}")  \\
        \textbf{Output} & reasoning: Chronicle of Malaysia gives an account of \underline{Malaysia} from January 1957 to 31 August 2007.\\
        & reasoning: Treason by the Book is a historical account of the Zeng Jing case which took place during the reign of the Yongzheng Emperor of \underline{Qing} China around 1730.\\
        & reasoning: Korea: A Walk Through the Land of Miracles is a book by Simon Winchester. He recounts his experience walking across \underline{South Korea}, from Jeju in the south to the DMZ in the north.\\
        & Final Answer: [\textquotesingle Chronicle of Malaysia\textquotesingle, \textquotesingle Korea: A Walk Through the Land of Miracles\textquotesingle, \textquotesingle Treason by the Book\textquotesingle] \correct{} \\
        \midrule
        \textbf{Input} & \texttt{\method{}-prompt}(query="Films about sisters shot in \underline{Paris}") \\
        \textbf{Output} & reasoning: Peppermint Soda is a 1977 French comedy-drama film directed by Diane Kurys. The film follows two teenage sisters over the course of the year 1963. The high school where the film takes place is the Lycée Jules-Ferry in \underline{Paris}, France.\\
        & Final Answer: [\textquotesingle Peppermint Soda\textquotesingle] \wrong{} // Partially correct (missing \textquotesingle Le Divorce\textquotesingle{} and \textquotesingle Two English Girls\textquotesingle)\\
        \bottomrule
        \\
    \end{tabular}
}
\caption{Qualitative examples showcasing the successes and failures of the \method{} prompting in the RAG task.
We use Gemini 1.5 Pro for the analysis.
In each example, we underline correct reasoning chains provided by the model.}
\label{tab:qual_examples_rag}
\end{table}

\begin{table}[ht]
\centering
\footnotesize
\resizebox{1.0\linewidth}{!}{
    \begin{tabular}{p{15mm}p{120mm}}
        \toprule
        \multicolumn{2}{c}{Task = {\bf SQL}} \\
        \midrule
        \multicolumn{2}{c}{Dataset = \textbf{Spider} 1M} \\
        \midrule
        \textbf{Input} & \texttt{\method{}-prompt}(query="Find the year which offers the largest number of courses.")  \\
        \textbf{Output} & Here's a step-by-step approach using the provided tables:\\\\ 
        & **1. Identify relevant data:**\\ 
        & We need to look at the "year" column in the "section" table, which lists the year each course section was offered.\\\\ 
        & **2. Count courses per year:**\\ 
        & Group the data by ``year'' and count the number of unique ``course\_id'' values for each year. This will give us the number of distinct courses offered in each year.\\\\ 
        & **3. Find the year with the maximum count:**\\ 
        & Identify the year with the highest count of distinct courses.\\\\ 
        & **Based on the data provided, here's the breakdown of courses per year:**\\\\ 
        & | Year | Number of Courses |\\ 
        & |---|---|\\ 
        & | 2001 | 29 |\\ 
        & | 2002 | 36 |\\ 
        & | 2003 | 38 |\\ 
        & | 2004 | 31 |\\ 
        & | 2005 | 31 |\\ 
        & | 2006 | 41 |\\ 
        & | 2007 | 40 |\\ 
        & | 2008 | 37 |\\ 
        & | 2009 | 35 |\\ 
        & | 2010 | 39 |\\
        & **The year 2006 has the largest number of courses offered (41).**\\\\ 
        &  Final Answer: [\textquotesingle 2006\textquotesingle] \correct{}\\ 
        \midrule
        \textbf{Input} & \texttt{\method{}-prompt}(query="List total amount of  invoice from Chicago, IL.")  \\
        \textbf{Output} & Here's a step-by-step approach using the provided tables:\\\\ 
        & **1. Identify relevant data:**\\
        & * **Invoices:** This table contains the billing city and total for each invoice.\\\\
        & **2. Filter invoices from Chicago, IL:**\\
        & * Scan through the ``billing\_city'' column and identify invoices where the city is ``Chicago'' and the state is ``IL''.\\ 
        & * **Based on the data provided, invoices 24, 92, 103, 158, 287, 310, 332, and 384 are from Chicago, IL.**\\\\ 
        & **3. Sum the total amounts:**\\
        & * Add up the "total" values for the invoices identified in step 2.\\ 
        & * **The total amounts for the invoices are 1.98, 1.98, 15.86, 8.91, 1.98, 7.96, 5.94, and 0.99, respectively.**\\ 
        & * **The sum of these amounts is 36.6.**\\\\
        & **4. Present the result:**\\
        & * The final output will be a single value representing the total amount of invoices from Chicago, IL.\\\\ 
        & Final Answer: [\textquotesingle 36.6\textquotesingle] \wrong{}\\
        \bottomrule
        \\
    \end{tabular}
}
\caption{Qualitative examples showcasing the successes and failures of the \method{} prompting in the SQL task.
We use Gemini 1.5 Pro for the analysis.
In SQL, the long-context model first outputs a reasoning chain in natural language ideally simulating the execution of the SQL query before arriving at the prediction.
}
\label{tab:qual_examples_sql}
\end{table}

\end{document}